\documentclass[lettersize,journal]{IEEEtran}
\usepackage{amsmath,amsfonts}
\usepackage{algorithmic}
\usepackage{algorithm}
\usepackage{array}
\usepackage[caption=false,font=normalsize,labelfont=sf,textfont=sf]{subfig}
\usepackage{textcomp}
\usepackage{stfloats}
\usepackage{url}
\usepackage{verbatim}
\usepackage{graphicx}
\usepackage{hyperref}
\usepackage{multirow}
\usepackage{multicol}
\usepackage{listings}
\usepackage{booktabs}
\usepackage{xcolor}
\usepackage{colortbl}
\usepackage[square, comma, sort&compress, numbers]{natbib}
\definecolor{fbApp}{HTML}{c8e7fa}
\definecolor{lightred}{RGB}{255, 230, 230}
\newcommand{\rrc}{\rowcolor{lightred}}
\begin{document}

\newlength{\logowidth}
\setlength{\logowidth}{10ex}
\newlength{\logospace}
\setlength{\logospace}{0.0em}

\title{%
	\makebox[\linewidth]{%
		\raisebox{-0.6\height}{\includegraphics[width=\logowidth]{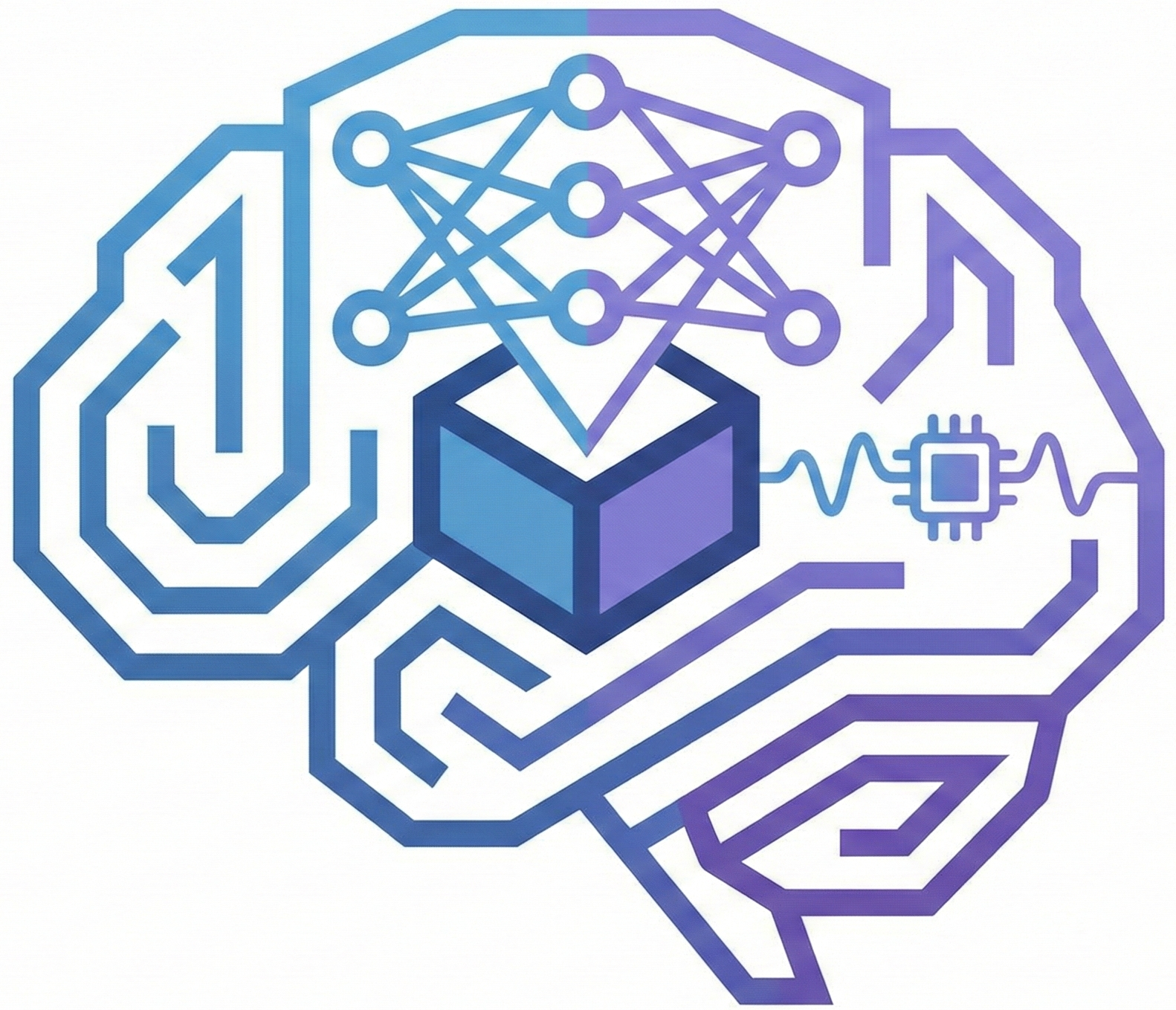}}%
		\hfill%
		\parbox[t]{0.83\linewidth}{\centering DeeperBrain: A Neuro-Grounded EEG Foundation Model Towards Universal BCI}%
		\hfill%
		\phantom{\raisebox{-0.5\height}{\includegraphics[width=\logowidth]{logo}}}%
	}%
}


\author{Jiquan Wang, Sha Zhao, Yangxuan Zhou, Yiming Kang, Shijian Li, Gang Pan,~\IEEEmembership{Senior Member,~IEEE}
\thanks{This work was supported in part by STI 2030 Major Projects under Grant 2021ZD0200400, in part by National Natural Science Foundation of China under Grant 62476240, in part by the Key Program of the Natural Science Foundation of Zhejiang Province, China, under Grant LZ24F020004, and in part by the Fundamental Research Funds for the Central Universities under Grant 226-2025-00122. \textit{(Jiquan Wang and Sha Zhao contributed equally to this work.) (Corresponding author: Sha Zhao.)}}
\thanks{Jiquan Wang is with the State Key Laboratory of Brain-machine Intelligence, Zhejiang University, Hangzhou 311121, Zhejiang, China (e-mail: wangjiquan@zju.edu.cn)}
\thanks{Sha Zhao, Yangxuan Zhou, Yiming Kang, and Shijian Li are with the State Key Laboratory of Brain-machine Intelligence, Zhejiang University, Hangzhou 311121, Zhejiang, China, and the College of Computer Science and Technology, Zhejiang University, Hangzhou 310013, Zhejiang, China (e-mail: szhao@zju.edu.cn; zyangxuan@zju.edu.cn; kangyiming@zju.edu.cn; shijianli@zju.edu.cn).}
\thanks{Gang Pan is with the State Key Laboratory of Brain-machine Intelligence, Zhejiang University, Hangzhou 311121, Zhejiang, China, the College of Computer Science and Technology, Zhejiang University, Hangzhou 310013, Zhejiang, China and the MOE Frontier Science Center for Brain Science and Brain-machine Integration, Hangzhou 310063, Zhejiang, China (e-mail: gpan@zju.edu.cn)}

}



\maketitle

\begin{abstract}
	Electroencephalography (EEG) foundation models hold significant promise for universal Brain-Computer Interfaces (BCIs). However, existing approaches often rely on  \textit{end-to-end fine-tuning} and exhibit limited efficacy under \textit{frozen-probing} protocols, lacking the intrinsic universality required for broad generalization. This limitation stems from adapting general-purpose sequence architectures that overlook the biophysical and dynamical principles of neural activity. To bridge this gap, we propose DeeperBrain, a neuro-grounded foundation model integrating domain-specific inductive biases into its model design and learning objectives. Architecturally, DeeperBrain incorporates a volume conduction-aware channel encoding to model spatial mixing via 3D geometry, and a neurodynamics-aware temporal encoding capturing slow adaptations using oscillatory and exponential bases. For pretraining, we introduce a dual-objective strategy combining Masked EEG Reconstruction (MER) for local fidelity and Neurodynamics Statistics Prediction (NSP). NSP enforces alignment with macroscopic brain states by predicting interpretable order parameters, including spectral power, functional connectivity, cross-frequency coupling, and dynamic complexity. Extensive experiments demonstrate that DeeperBrain achieves state-of-the-art or highly competitive performance under \textit{end-to-end fine-tuning}. Crucially, it maintains superior efficacy under a rigorous \textit{frozen-probing} protocol, verifying that embedding neuroscientific first principles endows learned representations with the intrinsic universality essential for universal BCI. The code will be publicly available.
\end{abstract}


\section{Introduction}
\label{sec:intro}

Brain-Computer Interfaces (BCIs) establish direct communication pathways between the central nervous system and external devices~\citep{schalk2004bci2000, wu2013convergence, zhang2019human}. Electroencephalography (EEG) serves as the predominant non-invasive modality for BCI systems, capable of capturing macroscopic cortical activity with millisecond-level temporal resolution, high portability, and cost-efficiency. Consequently, EEG has become the cornerstone of critical applications such as motor imagery~\citep{altaheri2021deep, dai2020hs}, affective state recognition~\citep{dadebayev2022eeg, gao2024multimodal, zhao2025wearable}, epilepsy diagnosis~\citep{ahmad2022eeg, yildiz2022unsupervised}, sleep staging~\citep{phan2021xsleepnet, wang2024generalizable, zhou2024personalized}, consciousness assessment~\citep{bai2021managing, zhao2024docter}, and depression detection~\citep{yasin2021eeg, wang2025m}. However, achieving universal BCI remains challenging, impeded by the multifaceted heterogeneity of neural data. At the signal level, recordings are inherently non-stationary and suffer from low signal-to-noise ratios, stemming from anatomical variability~\citep{saha2020intra} and pervasive artifacts like muscle activity and line noise~\citep{bjork1953electrical, mannan2018identification, uriguen2015eeg, jamil2021artifact}. Physically, diverse acquisition hardware employing varying channel counts and reference schemes induces inconsistent spatial alignments across datasets~\citep{acharya2019overview}. Compounding these discrepancies is the functional heterogeneity of BCI paradigms themselves, which recruit divergent neural mechanisms ranging from frequency-specific sensorimotor rhythms to complex, widespread network oscillations. These factors collectively prevent cross-scenario generalization, necessitating extensive adaptation to novel deployment contexts and hindering the realization of universal BCI systems.

Inspired by the success of large language and vision models~\citep{bommasani2021opportunities, min2023recent, awais2025foundation, xu2025lvlm}, the field has increasingly turned toward EEG foundation models to learn universal representations from vast unlabeled corpora~\citep{zhou2025brain, li2025foundation, gu2025foundation, lai2025simple, yang2025foundation}. These approaches typically employ self-supervised strategies like masked signal reconstruction. Despite their statistical capacity, these data-driven models rely on \textbf{\textit{end-to-end fine-tuning}} to adapt to downstream tasks yet frequently exhibit limited efficacy under a \textbf{\textit{frozen-probing}} protocol~\citep{lee2025are, xiong2025eeg, wu2025adabrain}. We argue that establishing robust performance in this frozen setting, where the pretrained backbone remains fixed, is a prerequisite for universal BCI. It serves as a critical benchmark for validating whether learned representations capture genuinely invariant neurophysiological mechanisms rather than scenario-specific statistics. We attribute the current deficiency to the reliance on generic sequence modeling assumptions. By prioritizing data-driven patterns while underutilizing the distinct biophysical and dynamical principles governing neural activity, existing architectures yield representations that lack intrinsic universality.

This limitation stems from neglecting the structured physical processes that define EEG. Two core principles shape this structure. First, \textbf{\textit{spatial organization}} is dictated by \textbf{\textit{volume conduction}}, where passive current spread creates spatially blurred, distance-dependent correlations among electrodes~\citep{nunez2006electric}. Second, \textbf{\textit{temporal organization}} exhibits a hierarchy spanning from millisecond-level oscillations to second-level cognitive adaptations~\citep{buzsaki2006rhythms, fairhall2001efficiency}. Furthermore, from a dynamical systems perspective, macroscopic brain function is governed by low-dimensional \textbf{\textit{order parameters}}~\citep{strogatz2024nonlinear, breakspear2017dynamic}. Key statistics like spectral power~\citep{buzsaki2006rhythms}, functional connectivity~\citep{varela2001brainweb}, cross-frequency coupling~\citep{canolty2006high}, and dynamic complexity~\citep{richman2000physiological} constitute the stable, interpretable features of brain states. Neglecting these priors yields neurobiologically agnostic representations, which necessitates extensive parameter updates to recover meaningful features, thereby hindering the emergence of the intrinsic universality required for robust \textit{frozen-probing}.

To bridge this gap, we propose to integrate neurophysiological principles directly into the foundation model design at both the architectural and objective levels. We introduce \textbf{DeeperBrain}, a neuro-grounded EEG foundation model that leverages domain-specific inductive biases to guide representation learning. Architecturally, DeeperBrain incorporates \textbf{a channel positional encoding that models volume conduction via 3D electrode geometry and a learnable spatial decay kernel}, alongside \textbf{a temporal positional encoding that uses oscillatory and exponential decay bases to capture slow modulations and adaptation}. For pretraining, we devise a dual-objective framework combining Masked EEG Reconstruction (MER) with \textbf{Neurodynamics Statistics Prediction (NSP)}. While MER preserves local signal fidelity, NSP enforces alignment with macroscopic brain states by predicting interpretable statistics derived from the full signal. By explicitly modeling these physical and dynamical constraints, DeeperBrain aligns representations with underlying neural dynamics. This fosters intrinsic universality, enabling robust performance even in the \textit{frozen-probing} setting without further adaptation.

Our main contributions are summarized as follows:
\begin{enumerate}
	\item \textbf{DeeperBrain: A Neuro-Grounded EEG Foundation Model.} We propose DeeperBrain, a foundation model that systematically incorporates neurophysiological principles into its model design and learning objective, facilitating generic representation learning across diverse EEG domains towards universal BCI.
	\item \textbf{Neurophysiologically Grounded Model Design.} We introduce two inductive biases aligned with EEG biophysics: (i) a volume conduction-aware channel positional encoding that models spatial mixing via 3D geometry, and (ii) a neurodynamics-aware temporal positional encoding that captures slow oscillations and adaptive dynamics via biologically plausible bases.
	\item \textbf{Neurodynamics-Guided Pretraining.} We introduce a dual-objective strategy optimizing both masked EEG reconstruction and neurodynamics statistics prediction. It ensures the learned representations capture both fine-grained waveform details and high-level dynamical order parameters essential for defining brain states.
	\item \textbf{Universal Representation.} Pretrained on 14 diverse datasets, DeeperBrain achieves state-of-the-art or competitive performance on a wide range of downstream tasks. Notably, it maintains superior performance even under the \textbf{\textit{frozen-probing}} protocol, verifying the universality and transferability of its learned features.
\end{enumerate}

\section{Related Work}

\subsection{Traditional EEG Decoding}
Electroencephalography (EEG) decoding aims to infer cognitive, affective, or pathological states from scalp-recorded electrical activity. Early methodologies predominantly relied on handcrafted features, including time-domain statistics, spectral power, and connectivity measures, coupled with classifiers such as support vector machines or linear discriminant analysis~\citep{mcfarland2006bci, lotte2007review, wu2014probabilistic, wang2023sparse}. These approaches necessitate extensive domain expertise and often exhibit limited generalization across diverse subjects and recording conditions.

The emergence of deep learning has shifted the field toward end-to-end representation learning~\citep{lecun2015deep}. Numerous studies now employ deep architectures to automate feature extraction from raw or minimally processed EEG~\citep{parvaneh2019cardiac, craik2019deep, al2021deep, sekkal2022automatic, yangpyhealth}. Convolutional Neural Networks (CNNs) have been widely adopted to capture local spatiotemporal patterns, with architectures such as DeepConvNet~\citep{schirrmeister2017deep}, C2CM~\citep{sakhavi2018c2cm}, EEGNet~\citep{lawhern2018eegnet}, and TSception~\citep{ding2022tsception} demonstrating efficacy in motor imagery, emotion recognition, and seizure detection~\citep{abdelhameed2021deep}. To model long-range temporal dependencies, Recurrent Neural Networks (RNNs), particularly Long Short-Term Memory (LSTM) networks, have been applied to tasks including sleep staging and BCI control~\citep{wang2018lstm, phan2019seqsleepnet}. Hybrid CNN-LSTM models further combine local feature extraction with sequential modeling for applications such as sleep staging and emotion recognition~\citep{supratak2017deepsleepnet, zhang2019cnnlstm, dar2020cnn, li2022motor, wang20232d}.

More recently, the Transformer architecture has been introduced to EEG analysis, utilizing self-attention to model global spatiotemporal interactions without fixed convolutional inductive biases~\citep{song2021transformer, liu2021transformers, du2022eeg, phan2022sleeptransformer}. Several works integrate CNNs and Transformers to jointly exploit local invariance and global context~\citep{song2022eeg, peh2022transformer, wang2023narcolepsy, zhou2024simplifying, wang2024caresleepnet}. Graph Neural Networks (GNNs) have also been explored to explicitly model inter-channel relationships using predefined or learned connectivity structures~\citep{jia2020graphsleepnet, jia2021multi, ding2023lggnet}. Despite success on specific benchmarks, these supervised methods face fundamental limitations: the high cost of acquiring labeled data and poor cross-dataset generalization due to variations in hardware, montage, and physiology. This has motivated the shift toward self-supervised pretraining.

\subsection{EEG Foundation Models}
Foundation models have revolutionized vision, language, and multimodal learning~\citep{bommasani2021opportunities, devlin2018bert, he2022masked, radford2021learning, achiam2023gpt, kirillov2023segment, videoworldsimulators2024}. Extending this paradigm to time series, general-purpose models have been proposed for domains including climate, finance, and traffic~\citep{eldele2021time, zhang2022self, woo2024unified, chen2024visionts, deng2024lpsgm}, with a subset targeting neural signals.

In the EEG domain, early self-supervised efforts focused on pretext tasks to extract invariant representations from unlabeled data~\citep{banville2021uncovering}. BENDER~\citep{kostas2021bendr} employs contrastive learning on generic EEG embeddings, while MAEEG~\citep{chien2022maeeg} utilizes a masked autoencoder framework for signal reconstruction. Subsequent works extended these concepts to intracranial recordings: BrainBERT~\citep{wang2022brainbert} reconstructs masked spectrograms of stereo-EEG, and Brant~\citep{zhang2023brant} introduces a Transformer-based model for long-term dependency modeling. Its extensions, Brant-2~\citep{yuan2024brant} and Brant-X~\citep{zhang2024brant}, further incorporate scalp EEG and multi-modal physiological alignment.

Recent research has increasingly targeted scalp EEG. BIOT~\citep{yang2023biot} employs a linear transformer for joint pretraining across diverse biosignals. EEG2Rep~\citep{foumani2024eeg2rep} predicts masked segments in a latent space, and LaBraM~\citep{jiang2024large} introduces neural token prediction for EEG patches. Neuro-GPT~\citep{cui2024neuro} utilizes a decoder-only architecture with autoregressive pretraining, while EEGPT~\citep{wang2024eegpt} employs a dual objective for universal feature extraction. Approaches such as NeuroLM~\citep{jiang2025neurolm}, UniMind~\citep{lu2025unimind}, and ELASTIQ~\citep{jiang2025elastiq} explore harnessing Large Language Models (LLMs) by treating EEG as a foreign language. Architecturally, CBraMod~\citep{wang2025cbramod} uses criss-cross attention with asymmetric positional encoding, EEGMamba~\citep{wang2025eegmamba} and CodeBrain~\citep{ma2025codebrain} leverage state space models for spatiotemporal dependencies, and LUNA~\citep{doner2025luna} unifies diverse electrode layouts via learned queries. Other innovations include mixture-of-experts in NeurIPT~\citep{fang2025neuript}, structured sparse attention in CSBrain~\citep{zhou2025csbrain}, discrete tokenization in BrainOmni~\citep{xiao2025brainomni}, and massive-scale pretraining in REVE~\citep{ouahidi2025reve}.

While these models improve transferability, they predominantly adapt generic architectures from vision or language. Lacking explicit biophysical constraints, they rely heavily on \textit{end-to-end fine-tuning} and typically struggle under \textit{frozen-probing} protocols~\citep{lee2025are, xiong2025eeg, wu2025adabrain}.

\begin{figure*}[tb]
	\centering
	\small
	\includegraphics[width=\textwidth]{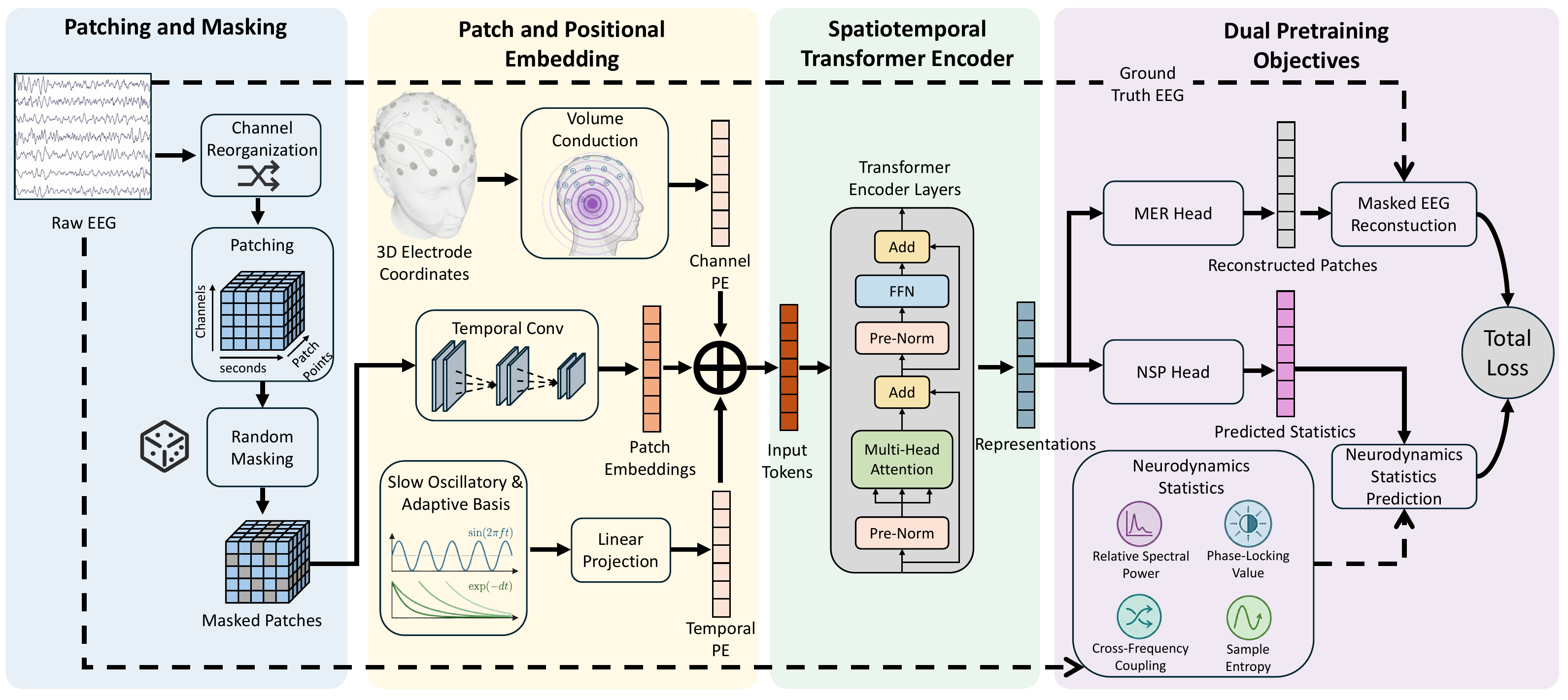}
	\caption{DeeperBrain overview.}
	\label{fig:model}
\end{figure*}

\section{Method}
\label{sec:method}

\subsection{Overview}
\label{subsec:overview}

DeeperBrain is a neuro-grounded foundation model designed to learn universal representations from unlabeled EEG. As illustrated in Fig.~\ref{fig:model}, the pipeline begins by partitioning the channel-reorganized EEG signal into non-overlapping spatiotemporal patches. A random subset of these patches is masked, while the visible ones are projected into dense tokens via a temporal convolutional backbone. To inject biophysical inductive biases, we augment these tokens with two domain-specific positional encodings: a channel encoding that models volume conduction using 3D electrode geometry, and a temporal encoding that captures slow oscillatory and adaptive dynamics. These tokens are then processed by a Transformer encoder to model global dependencies. Finally, DeeperBrain is optimized via a dual-objective strategy: Masked EEG Reconstruction (MER) to preserve local signal fidelity, and Neurodynamics Statistics Prediction (NSP) to enforce alignment with macroscopic neural principles, including spectral organization and functional connectivity.

\subsection{Patching and Masking Strategy}
\label{subsec:patching}

To enhance robustness to variable electrode montages and promote generalization across heterogeneous datasets, we introduce a stochastic \textit{channel reorganization} step. Let the raw EEG be denoted as $\widetilde{\mathbf{X}} \in \mathbb{R}^{\widetilde{C} \times T}$, where $\widetilde{C}$ is the original channel count and $T$ represents time samples. For each training instance, we randomly permute the channel order to eliminate positional bias and select a subset of size $C \in [1, \widetilde{C}]$ to simulate variable-density recordings. This yields the reorganized signal:
\begin{equation}
	\small
	\mathbf{X} = \texttt{channel\_reorganization}(\widetilde{\mathbf{X}}) \in \mathbb{R}^{C \times T}.
\end{equation}
By treating $C$ as a stochastic variable, the model learns to handle arbitrary channel configurations during both pretraining and inference.

The reorganized signal $\mathbf{X}$ is partitioned into non-overlapping one-second segments. Given a sampling frequency $f_s$, each segment contains $P = f_s$ points. We reshape the signal into a 3D tensor:
\begin{equation}
	\small
	\mathbf{X}_{\text{patch}} \in \mathbb{R}^{C \times S \times P},
\end{equation}
where $S = T / P$ denotes the number of temporal segments. Each vector $\mathbf{X}_{c,s,:} \in \mathbb{R}^{P}$ constitutes a $(channel \times second)$ patch, serving as the atomic unit for self-supervised learning.

During pretraining, we employ a random patch-wise masking strategy. For each recording, a fraction $\mu$ (default $\mu = 0.5$) of the total $C \times S$ patches is selected uniformly across channels and time. Masked patches are replaced with a fixed zero vector $\mathbf{0} \in \mathbb{R}^{P}$ in the raw voltage space rather than a learnable token, ensuring compatibility with the temporal convolutional embedding. This operation is defined by a binary mask $\mathbf{M} \in \{0,1\}^{C \times S}$, where $\mathbf{M}_{c,s} = 1$ indicates a masked patch. By masking entire $(channel \times second)$ units, we prevent the model from relying on local correlations and force it to reconstruct missing content using global spatiotemporal context, thereby preserving the integrity of unmasked neural dynamics.

\subsection{Patch Embedding via Temporal Convolution}
\label{subsec:patch_embedding}

To transform raw voltage patches into dense, semantically rich embeddings, we utilize a lightweight temporal convolutional network. This backbone leverages the insight that each $(channel \times second)$ patch contains local temporal structures, such as oscillatory bursts and transient events, effectively captured by 1D convolutions. For efficient implementation, we reshape the masked input $\mathbf{X}_{\text{mask}} \in \mathbb{R}^{C \times S \times P}$ into $\mathbb{R}^{1 \times (C S) \times P}$, treating the spatiotemporal grid as a set of independent 1D signals.

The embedding architecture comprises three sequential convolutional blocks designed to extract multi-scale features:
\begin{equation}
	\small
	\mathbf{E}_{\text{time}} = \text{ConvBlock}_3 \circ \text{ConvBlock}_2 \circ \text{ConvBlock}_1(\mathbf{X}_{\text{mask}}).
\end{equation}
Each $\text{ConvBlock}_i$ consists of a 2D convolution with kernel size $(1, K_i)$ and stride $(1, s_i)$, followed by Group Normalization (5 groups) and GELU activation. The first block employs a large kernel ($K_1 = 49$, $s_1 = 25$) to capture long-range dependencies and reduce sequence length, effectively mimicking delta–theta bandpass filtering. Subsequent blocks use small kernels ($K_2 = K_3 = 3$) with unit stride to refine local structures like alpha or beta bursts without further downsampling. This hierarchical design extracts both coarse and fine temporal features while maintaining computational efficiency.

The resulting feature tensor $\mathbf{E}_{\text{time}} \in \mathbb{R}^{F \times (C S) \times L}$, with $F = 25$ output channels and length $L$, is flattened to yield the final patch embeddings:
\begin{equation}
	\small
	\mathbf{E} \in \mathbb{R}^{C \times S \times D},
\end{equation}
where $D = F \times L$ is the embedding dimension. These embeddings provide the initial token sequence for the Transformer encoder, encoding rich local temporal context while preserving the original spatiotemporal organization.

\subsection{Neurophysiologically Grounded Positional Encodings}
\label{subsec:positional_encodings}
Standard positional encodings in vision or language models assume either fixed geometric priors (e.g., 2D grid in ViT~\citep{dosovitskiy2020image}) or purely learnable embeddings. While sufficient for many domains, EEG is governed by distinct biophysical laws. Explicitly modeling these principles can provide stronger inductive biases than generic learnable embeddings. To address this, we introduce two \textit{neurophysiologically grounded positional encodings} that explicitly embed the physics of EEG generation into the model architecture.

\textbf{Channel Positional Encoding: Modeling Volume Conduction.}  
Scalp EEG potentials arise from the passive spread of electrical currents through the head’s conductive tissues, a process known as \textit{volume conduction}~\citep{nunez2006electric}. Under the quasi-static approximation of Maxwell’s equations, the potential $\phi(\mathbf{r})$ at scalp location $\mathbf{r}$ due to a current dipole $\mathbf{q}$ at source location $\mathbf{r}_s$ is:
\begin{equation}
	\small
	\phi(\mathbf{r}) = \frac{1}{4\pi\sigma} \frac{\mathbf{q} \cdot (\mathbf{r} - \mathbf{r}_s)}{\|\mathbf{r} - \mathbf{r}_s\|^3},
\end{equation}
where $\sigma$ is the effective conductivity of head tissues. This implies that signal similarity decays with physical distance between electrodes, inducing strong spatial correlations among nearby sensors.

To model this, we leverage 3D electrode coordinates $\mathbf{P}_{\text{ch}} \in \mathbb{R}^{C \times 3}$ in a standardized anatomical space (e.g., MNI, in meters). Pairwise Euclidean distances are computed as $\mathbf{D}_{ij} = \|\mathbf{p}_i - \mathbf{p}_j\|_2$. Inspired by the exponential decay observed in volume conduction models~\citep{grech2008review}, we define a learnable spatial kernel:
\begin{equation}
	\small
	\mathbf{K}_{ij} = \exp\left(-\frac{\mathbf{D}_{ij}}{\tau}\right), \quad \tau = \text{softplus}(\alpha) + \epsilon,
\end{equation}
where $\alpha$ is a learnable parameter controlling the decay scale $\tau > 0$, initialized to correspond to a typical inter-electrode distance (8 cm).

From a biophysical perspective, EEG can be modeled as a linear mixture $\mathbf{x} = \mathbf{A} \mathbf{s} + \boldsymbol{\epsilon}$, where $\mathbf{A} \in \mathbb{R}^{C \times Q}$ is the lead field matrix and $\mathbf{s}$ denotes source activity. Under isotropic and uncorrelated sources ($\mathbb{E}[\mathbf{s}\mathbf{s}^\top] = \mathbf{I}$), the covariance of observed signals satisfies $\Sigma_{ij} \propto \exp(-\|\mathbf{p}_i - \mathbf{p}_j\| / \tau)$~\citep{nunez2006electric}. Our kernel $\mathbf{K}_{ij}$ thus approximates this expected spatial correlation structure. To ensure stable aggregation, we normalize each row of the kernel to form a convex combination:
\begin{equation}
	\small
	\bar{\mathbf{K}}_{ij} = \frac{\mathbf{K}_{ij}}{\sum_{k=1}^C \mathbf{K}_{ik} + \epsilon}.
\end{equation}
The smoothing operation
\begin{equation}
	\small
	\tilde{\mathbf{p}}_i = \sum_{j=1}^C \bar{\mathbf{K}}_{ij} \mathbf{p}_j
\end{equation}
embeds each electrode in a \textit{functional coordinate system}, where its effective position is a weighted average of all electrodes’ anatomical locations, with weights determined by physical proximity. This mimics the volume-conducted mixing of neural sources and produces a spatially contextualized position vector for each channel. Finally, $\tilde{\mathbf{p}}_i$ is projected to the embedding dimension $D$ via a linear layer and added to the patch embeddings, endowing the model with an inductive bias aligned with EEG physics.

\textbf{Temporal Positional Encoding: Capturing Slow Modulations and Adaptation.}  
While temporal convolutional patch embeddings capture fine-grained oscillations, higher-order brain states involve slower modulations spanning seconds to minutes, including alpha power fluctuations and adaptation to sustained stimuli~\citep{buzsaki2006rhythms, fairhall2001efficiency}. These dynamics provide critical temporal context. However, generic positional encodings, including standard sinusoidal and learnable embeddings, treat time indices as symmetric dimensions or arbitrary markers. They lack inductive biases for the "arrow of time," failing to capture the dissipative nature of neural adaptation. To model this cross-second structure, we design a temporal positional encoding operating at \textit{second-level} granularity that combines two biologically inspired bases:

\textit{Slow oscillatory basis}: We model periodic modulations in the range of 0.01–0.5 Hz, corresponding to time scales from 2 seconds (0.5 Hz) to 100 seconds (0.01 Hz). For each fixed frequency $f_k$ in this range, we include sine and cosine components:
\begin{equation}
	\small
	\psi_k^{\text{osc}}(t) = \left[ \sin\left(2\pi f_k t\right),\ \cos\left(2\pi f_k t\right) \right],
\end{equation}
where $t \in \{1, \dots, S\}$ denotes absolute time in seconds. This band captures slow cognitive and experimental-state fluctuations, such as block-wise attentional engagement or sleep stage transitions, that persist even after high-pass filtering of raw EEG. Frequencies are pre-defined on a logarithmic scale to prioritize behaviorally relevant low frequencies, providing a stable inductive bias without learnable parameters.

\textit{Adaptive decay basis}: To model neural adaptation or temporal forgetting over tens of seconds, we include exponential decay functions with fixed rates $d_m$:
\begin{equation}
	\small
	\psi_m^{\text{dec}}(t) = \exp\left(-d_m t\right).
\end{equation}
This is grounded in the observation that neural responses to sustained input often follow $r(t) \propto e^{-t/\tau}$~\citep{fairhall2001efficiency}, where $\tau = 1/d_m$ is the adaptation time constant. The decay rates are pre-defined to span $\tau \in [1, 100]$ seconds, covering sensory adaptation (seconds) to cognitive maintenance (tens of seconds), and are held fixed during training.

The full temporal feature vector for time step $t$ is the concatenation:
\begin{equation}
	\small
	\boldsymbol{\psi}(t) = \left[ \psi_1^{\text{osc}}(t), \dots, \psi_{K}^{\text{osc}}(t), \psi_1^{\text{dec}}(t), \dots, \psi_{M}^{\text{dec}}(t) \right] \in \mathbb{R}^{2K + M}.
\end{equation}
This vector is linearly projected to $D$ dimensions and broadcast across all channels. The fixed bases provide a principled, non-adaptive temporal prior that complements the fast temporal features extracted by the convolutional backbone, enabling the Transformer to jointly model dynamics across multiple time scales. The resulting encoding is both \textit{biologically plausible} and \textit{sufficiently expressive} to capture the hierarchical temporal organization of real EEG signals.

\subsection{Spatiotemporal Transformer Encoder}
\label{subsec:transformer}

The embedded spatiotemporal tokens $\mathbf{Z}^{(0)} \in \mathbb{R}^{N \times D}$, with $N = C \times S$ patches and embedding dimension $D$, are processed by a stack of $L$ Transformer encoder layers. While we adopt the standard pre-normalization architecture~\citep{wang2019learning}, its suitability for EEG stems from a fundamental alignment with the brain’s own computational principles: neural processing is inherently \textit{distributed} (across regions) and \textit{dynamic} (across time scales). The self-attention mechanism naturally models such long-range, flexible interactions, mirroring how distant brain areas coordinate via phase synchronization or cross-frequency coupling during cognition.

Formally, each layer follows the pre-normalization scheme~\citep{wang2019learning}:
\begin{align}
	\small
	\widetilde{\mathbf{Z}} &= \mathbf{Z}^{(l-1)} + \text{MHA}\big( \text{LN}(\mathbf{Z}^{(l-1)}) \big), \\
	\mathbf{Z}^{(l)} &= \widetilde{\mathbf{Z}} + \text{FFN}\big( \text{LN}(\widetilde{\mathbf{Z}}) \big),
\end{align}
where MHA denotes multi-head self-attention and FFN a two-layer feedforward network with GELU activation. The attention operation enables each $(channel, second)$ token to dynamically integrate information from all other spatiotemporal locations, effectively simulating functional connectivity across the cortical surface and over time. For instance, the model can learn to link frontal theta oscillations (associated with working memory maintenance) with parietal gamma bursts (linked to sensory binding), recapitulating known neurocognitive pathways.

The final output $\mathbf{H} = \mathbf{Z}^{(L)} \in \mathbb{R}^{N \times D}$ thus yields a contextualized representation where each token encodes not only local neural dynamics (from the temporal convolutional embedding) but also its role within the global brain network. This joint modeling of space and time provides the inductive bias necessary for universal EEG decoding, without requiring explicit graph priors or fixed connectivity assumptions.

\subsection{Masked EEG Reconstruction (MER)}
\label{subsec:mer}

Preserving fine-grained temporal fidelity is critical for identifying transient neural events, such as epileptic spikes, sleep spindles, and muscle artifacts, which are essential for clinical and cognitive decoding. Given the susceptibility of raw EEG to non-stationary noise and high-amplitude transients, we employ a masked signal reconstruction objective to recover local waveform details from global spatiotemporal context.

The contextualized token sequence $\mathbf{H} \in \mathbb{R}^{N \times D}$ ($N = C \times S$) from the Transformer encoder is mapped back to the raw voltage space via a lightweight reconstruction head. This head comprises a single linear layer without activation:
\begin{equation}
	\small
	\hat{\mathbf{X}}_{\text{flat}} = \mathbf{H} \mathbf{W}_{\text{recon}}^\top + \mathbf{b}_{\text{recon}} \in \mathbb{R}^{N \times P},
\end{equation}
where $\mathbf{W}_{\text{recon}} \in \mathbb{R}^{P \times D}$ and $\mathbf{b}_{\text{recon}} \in \mathbb{R}^{P}$ are learnable parameters, and $P$ denotes the number of time samples per patch. The output is reshaped to restore the original spatiotemporal structure:
\begin{equation}
	\small
	\hat{\mathbf{X}} \in \mathbb{R}^{C \times S \times P}.
\end{equation}

The reconstruction loss is calculated exclusively on masked patches to prevent trivial identity mapping. Let $\mathcal{M} = \{(c,s) \mid \mathbf{M}_{c,s} = 1\}$ represent the set of masked indices, with $\mathbf{M} \in \{0,1\}^{C \times S}$ as the binary mask tensor. We utilize the Smooth L1 Loss (Huber loss)~\citep{huber1992robust}:
\begin{align}
	\small
	\mathcal{L}_{\text{MER}} &= \frac{1}{|\mathcal{M}| \cdot P} \sum_{(c,s) \in \mathcal{M}} \sum_{p=1}^{P} \rho\left( \hat{\mathbf{X}}_{c,s,p} - \mathbf{X}_{c,s,p} \right), \\
	\rho(e) &= 
	\begin{cases}
		\frac{1}{2}e^2 & \text{if } |e| < \beta, \\
		\beta(|e| - \frac{1}{2}\beta) & \text{otherwise},
	\end{cases}
\end{align}
where $\beta = 1.0$. This loss function balances mean squared error (MSE) for small residuals, enabling precise waveform recovery, with mean absolute error (MAE) for large residuals, ensuring robustness against outliers such as EMG or EOG artifacts commonly found in real-world EEG. Optimizing $\mathcal{L}_{\text{MER}}$ forces the model to learn representations grounded in the physical signal, capturing essential local details through context.

\subsection{Neurodynamics Statistics Prediction (NSP)}
\label{subsec:nsp}

While Masked EEG Reconstruction (MER) ensures local signal fidelity, it does not explicitly enforce the modeling of multi-scale dynamical structures defining brain states. To bridge this gap, we introduce \textit{Neurodynamics Statistics Prediction} (NSP), a self-supervision paradigm grounded in \textit{dynamical systems theory}~\citep{strogatz2024nonlinear, breakspear2017dynamic}. We conceptualize the brain as a high-dimensional nonlinear system whose macroscopic behavior is governed by a low-dimensional set of \textit{order parameters}, representing stable, emergent collective variables~\citep{buzsaki2010neural}.

Formally, let the latent neural state $\mathbf{z}(t) \in \mathbb{R}^d$ evolve according to:
\begin{equation}
	\small
	\frac{d\mathbf{z}(t)}{dt} = \mathbf{F}(\mathbf{z}(t), \mathbf{u}(t), t),
\end{equation}
where $\mathbf{F}$ is a nonlinear vector field and $\mathbf{u}(t)$ denotes external inputs. The observed EEG is a noisy projection:
\begin{equation}
	\small
	\mathbf{x}(t) = \mathcal{H}(\mathbf{z}(t)) + \boldsymbol{\epsilon}(t),
\end{equation}
with observation function $\mathcal{H}(\cdot)$ and noise $\boldsymbol{\epsilon}(t)$. While $\mathbf{z}(t)$ is high-dimensional, its emergent dynamics are characterized by macroscopic order parameters $\mathbf{y}(t) = \Phi(\mathbf{z}(t))$, where $\Phi$ extracts stable features robust to microscopic perturbations~\citep{buzsaki2010neural}.

Since $\mathbf{z}(t)$ is unobservable, we approximate $\mathbf{y}(t)$ directly from the EEG signal via a mapping $\Psi(\mathbf{x}(t)) \approx \mathbf{y}(t)$. In this work, $\Psi(\cdot)$ computes four neurophysiologically grounded statistics serving as proxies for canonical order parameters:

\textbf{1. Relative Spectral Power: Local Oscillatory States.} Spectral power distribution reflects the attractor landscape, with specific frequency bands linked to cognitive states~\citep{buzsaki2006rhythms}. We compute the Power Spectral Density (PSD) via real FFT:
\begin{equation}
	\small
	\text{PSD}(f) = \left| \sum_{t=0}^{P-1} x[t] e^{-j 2\pi f t / f_s} \right|^2,
\end{equation}
where $f_s$ is the sampling rate. Power is integrated over five canonical bands (delta to gamma) and normalized to a probability simplex:
\begin{equation}
	\small
	p_k = \frac{\int_{f \in \text{band}_k} \text{PSD}(f) \, df}{\sum_{k'} \int_{f \in \text{band}_{k'}} \text{PSD}(f) \, df}, \quad k = 1,\dots,5.
\end{equation}
This 5D vector quantifies the oscillatory order parameter.

\textbf{2. Phase-Locking Value (PLV): Functional Integration.} Phase synchronization measures large-scale functional integration~\citep{varela2001brainweb}. For each band, we extract the analytic signal phase $\phi_c(t)$ via the Hilbert transform. The PLV between channels $i$ and $j$ is:
\begin{equation}
	\small
	\text{PLV}_{ij} = \left| \frac{1}{P} \sum_{t=1}^{P} e^{j(\phi_i(t) - \phi_j(t))} \right|.
\end{equation}
We summarize the $C \times C$ PLV matrix by the mean and standard deviation of its upper-triangular elements, yielding a 10D vector capturing global synchronization strength and variability.

\textbf{3. Cross-Frequency Coupling (CFC): Hierarchical Coordination.} CFC encodes coordination across temporal scales, such as theta--gamma interactions~\citep{canolty2006high}. For canonical pairs (e.g., theta--gamma), we extract the low-frequency phase $\phi_{\text{low}}(t)$ and high-frequency amplitude $A_{\text{high}}(t)$, computing the modulation strength:
\begin{equation}
	\small
	\text{CFC} = \frac{1}{P} \sum_{t=1}^{P} \cos(\phi_{\text{low}}(t)) \cdot A_{\text{high}}(t).
\end{equation}
This 3D vector quantifies cross-scale coupling per patch.

\textbf{4. Sample Entropy: Dynamic Complexity.} Signal complexity reflects the richness of neural dynamics~\citep{richman2000physiological}. Sample entropy quantifies regularity:
\begin{equation}
	\small
	\text{SampEn}(m, r) = -\log \left( \frac{C^{m+1}(r)}{C^m(r)} \right),
\end{equation}
where $C^m(r)$ is the fraction of similar $m$-point templates within tolerance $r$. With $m=2$ and $r = 0.2 \times \text{std}(x)$, this yields a 1D measure of unpredictability.

These statistics form the target tensor $\mathbf{Y}_{\text{NS}} \in \mathbb{R}^{C \times S \times 19}$. The model infers these properties from the masked input $\mathbf{X}_{\text{mask}}$ via a linear prediction head:
\begin{equation}
	\small
	\hat{\mathbf{Y}}_{\text{NS}} = \mathbf{H} \mathbf{W}_{\text{NSP}}^\top + \mathbf{b}_{\text{NSP}} \in \mathbb{R}^{N \times 19}.
\end{equation}
The NSP loss is computed using Smooth L1 Loss:
\begin{equation}
	\small
	\mathcal{L}_{\text{NSP}} = \frac{1}{C S \cdot 19} \sum_{c,s,k} \rho\left( \hat{\mathbf{Y}}_{\text{NS},c,s,k} - \mathbf{Y}_{\text{NS},c,s,k} \right).
\end{equation}
By targeting these universal principles, NSP encourages the learning of representations invariant to subject-specific conditions~\citep{arjovsky2019invariant}. The total pretraining objective combines NSP with MER:
\begin{equation}
	\small
	\mathcal{L} = \lambda_{\text{MER}} \mathcal{L}_{\text{MER}} + \lambda_{\text{NSP}} \mathcal{L}_{\text{NSP}},
\end{equation}
where $\lambda_{\text{MER}}$ and $\lambda_{\text{NSP}}$ are the coefficients of two objectives. This dual approach ensures DeeperBrain captures both physical signal fidelity and interpretable neurodynamical mechanisms.

\begin{table*}[!tb]
	\centering
	\scriptsize
	\caption{Overview of pre-training datasets. }
	\begin{tabular}{lrrrrr} 
		\toprule
		\textbf{Datasets}&\textbf{Rate}&\textbf{\# Channels}&\textbf{Duration}&\textbf{\# Samples}&\textbf{Total Duration}\\
		\midrule
		TUEG (clean subset)~\citep{obeid2016temple}&256Hz&19&30s&1,109,545&9,246.2h\\
		PhysioNet 2018~\citep{Goldberger2000PhysioBankPA, ghassemi2018you}&200Hz&6&30s&872,089&7,267.4h\\
		Raw EEG Data~\citep{raweegdata} &256Hz&64&5s&30,978&42.7h\\
		Siena Scalp EEG Database~\citep{Goldberger2000PhysioBankPA, detti2020eeg}&256Hz&27&10s&50,749&140.8h\\
		ds006171~\citep{melcon2024perception, ds006171}&1024Hz&128&5s&29,397&40.8h\\
		ds006317~\citep{zhang2024chisco, ds006317}&1000Hz&109&5s&44,708&62.1h\\
		ds006367~\citep{ds006367}&1000Hz&28&8s&29,929&66.5h\\
		ds006370~\citep{ds006370}&1000Hz&28&5s&67,953&94.4h\\
		ds006437~\citep{ds006437}&256Hz&64&5s&12,069&16.8h\\
		ds006446~\citep{nashiro2024heart,kim2025electroencephalography,ds006446}&2048Hz&64&5s&12,970&18.0h\\
		ds006466~\citep{nashiro2024heart,kim2025electroencephalography,ds006466}&1000Hz&64&5s&94,392&131.1h\\
		ds006480~\citep{shou2020whole,ds006480}&1000Hz&64&5s&51,572&71.6h\\
		ds006525~\citep{ds006525}&250Hz&128&5s&4,080&5.7h\\
		ds006547~\citep{ds006547}&500Hz&63&5s&28,222&39.2h\\
		\bottomrule
	\end{tabular}
	\label{tab:pre-train}
\end{table*}

\begin{table*}[!tb]
	\centering
	\scriptsize
	\setlength{\tabcolsep}{3pt}
	\caption{Overview of downstream BCI tasks and datasets.}
	\begin{tabular}{llrrrrclll} 
		\toprule
		\textbf{BCI Tasks}&\textbf{Datasets}&\textbf{Rate}&\textbf{\# Channels}&\textbf{Duration}&\textbf{\# Samples}&\textbf{Label}&\textbf{Train}&\textbf{Val}&\textbf{Test}\\
		\midrule
		Emotion Recognition&FACED~\citep{chen2023large}&250Hz&32&10s&10,332&9-class&Sub 1-80&Sub 81-100&Sub 101-123\\
		&SEED-V~\citep{liu2021comparing}&1000Hz&62&1s&117,744&5-class&\multicolumn{3}{c}{Trials per session split 5:5:5}\\
		&SEED-VII~\citep{jiang2024seed}&1000Hz&62&5s&55,561&7-class&\multicolumn{3}{c}{Trials per session split 10:5:5}\\
		Motor Imagery Classification&PhysioNet-MI~\citep{goldberger2000physiobank, schalk2004bci2000}&160Hz&64&4s&9,837&4-class&Sub 1-70&Sub 71-89&Sub 90-109\\
		&BCIC-IV-2a~\citep{brunner2008bci}&250Hz&22&4s&5,088&4-class&Sub 1-5&Sub 6-7&Sub 8-9\\
		&SHU-MI~\citep{ma2022large}&250Hz&32&4s&11,988&2-class&Sub 1-15&Sub 16-20&Sub 21-25\\
		Sleep Staging&ISRUC~\citep{khalighi2016isruc}&200Hz&6&30s&89,240&5-class&Sub 1-80&Sub 81-90&Sub 91-100\\
		Seizure Detection&CHB-MIT~\citep{goldberger2000physiobank, shoeb2009application}&256Hz&16&10s&326,993&2-class&Sub 1-19&Sub 20-21&Sub 22-23\\
		Imagined Speech Classification&BCIC2020-3~\citep{jeong20222020}&256Hz&64&3s&6,000&5-class&\multicolumn{3}{c}{Predefined training, validation, and test}\\
		Mental Disorder Diagnosis&MODMA~\citep{cai2022multi}&250Hz&128&15s&1,066&2-class&15MDD+18NC&4MDD+5NC& 5MDD+6NC.\\
		Vigilance Estimation&SEED-VIG~\citep{min2017driver}&200Hz&17&8s&20,355&regression&Sub 1-15&Sub 16-19&Sub 20-23\\
		Workload Estimation&MentalArithmetic~\citep{goldberger2000physiobank, zyma2019electroencephalograms}&500Hz&20&5s&1,707&2-class&Sub 1-28&Sub 29-32&Sub 33-36\\
		\bottomrule
	\end{tabular}
	\label{tab:downstream}
\end{table*}

\subsection{Task-specific Prediction Head}
\label{subsec:classifier}

To evaluate the learned representations on downstream tasks, we attach a task-specific prediction head to the Transformer output $\mathbf{H} \in \mathbb{R}^{N \times D}$ ($N = C \times S$). This head maps the spatiotemporal token sequence to target outputs, employing either global pooling or full-sequence flattening depending on the task's temporal structure.

We assess performance under two protocols: (i) \textit{end-to-end fine-tuning} of the entire model, and (ii) \textit{frozen-probing} with the backbone frozen. The optimization objective is selected based on the task type. For classification, we minimize the cross-entropy loss. For regression tasks, we employ the Mean Squared Error (MSE) loss.
This evaluation strategy enables a direct assessment of representation quality: strong performance in the frozen setting indicates that DeeperBrain learns universal features without relying on task-specific fine-tuning.

\begin{table*}[tb]
	\scriptsize
	\setlength{\tabcolsep}{5pt}
	\centering
	\caption{Performance comparison with existing methods under end-to-end fine-tuning. }
	\begin{tabular}{lccccccccc} 
		\toprule
		&\multicolumn{3}{c}{\textbf{FACED, 9-class}}&\multicolumn{3}{c}{\textbf{SEED-V, 5-class}}&\multicolumn{3}{c}{\textbf{SEED-VII, 7-class}}\\
		\cmidrule(lr){2-4}\cmidrule(lr){5-7}\cmidrule(lr){8-10}
		\textbf{Methods}&\textbf{Bal. Acc.}&\textbf{Kappa}&\textbf{Weighted F1}&\textbf{Bal. Acc.}&\textbf{Kappa}&\textbf{Weighted F1}&\textbf{Bal. Acc.}&\textbf{Kappa}&\textbf{Weighted F1}\\
		\midrule
		EEGNet    
		&40.90 $\pm$ 1.22&33.42 $\pm$ 2.51&41.24 $\pm$ 1.41
		&29.61 $\pm$ 1.02&10.06 $\pm$ 1.43&27.49 $\pm$ 0.98
		&25.28 $\pm$ 0.33&12.85 $\pm$ 0.44&25.36 $\pm$ 0.36\\
		EEGConformer   
		&45.59 $\pm$ 1.25&38.58 $\pm$ 1.86&45.14 $\pm$ 1.07
		&35.37 $\pm$ 1.12&17.72 $\pm$ 1.74&34.87 $\pm$ 1.36
		&28.75 $\pm$ 0.28&17.47 $\pm$ 0.30&28.02 $\pm$ 0.28\\
		\midrule
		LaBraM  
		&52.73 $\pm$ 1.07&46.98 $\pm$ 1.88&52.88 $\pm$ 1.02
		&39.76 $\pm$ 1.38&23.86 $\pm$ 2.09&39.74 $\pm$ 1.11
		&\underline{33.28} $\pm$ 0.33&\underline{22.85} $\pm$ 0.44&\underline{33.36} $\pm$ 0.36\\
		CBraMod  
		&55.09 $\pm$ 0.89&50.41 $\pm$ 1.22&56.18 $\pm$ 0.93
		&\underline{40.91} $\pm$ 0.97&25.69 $\pm$ 1.43&41.01 $\pm$ 1.08
		&32.47 $\pm$ 0.36&21.47 $\pm$ 0.38&32.77 $\pm$ 0.32\\
		CSBrain  
		&\underline{57.52} $\pm$ 0.42&\underline{52.04} $\pm$ 0.36&\underline{57.96} $\pm$ 0.31
		&\textbf{41.97} $\pm$ 0.33&\textbf{27.85} $\pm$ 0.34&\textbf{42.80} $\pm$ 0.23
		&32.44 $\pm$ 0.38&21.45 $\pm$ 0.49&32.74 $\pm$ 0.47\\
		REVE
		&56.46 $\pm$ 1.64&50.80 $\pm$ 1.91&56.59$\pm$ 1.72
		&40.48 $\pm$ 0.23&26.22 $\pm$ 0.21&41.45$\pm$ 0.17
		&32.64 $\pm$ 0.34&21.72 $\pm$ 0.37&33.02$\pm$ 0.31\\
		\midrule
		\rrc DeeperBrain  
		&\textbf{60.32} $\pm$ 0.34&\textbf{54.99} $\pm$ 0.36&\textbf{60.58} $\pm$ 0.33
		&40.68 $\pm$ 0.31&\underline{26.43} $\pm$ 0.36&\underline{41.54} $\pm$ 0.25
		&\textbf{33.60} $\pm$ 0.25&\textbf{22.86} $\pm$ 0.28&\textbf{33.95} $\pm$ 0.23\\
		\midrule
		&\multicolumn{3}{c}{\textbf{PhysioNet-MI, 4-class}}&\multicolumn{3}{c}{\textbf{BCIC-IV-2a, 4-class}}&\multicolumn{3}{c}{\textbf{SHU-MI, 2-class}}\\
		\cmidrule(lr){2-4}\cmidrule(lr){5-7}\cmidrule(lr){8-10}
		\textbf{Methods}&\textbf{Bal. Acc.}&\textbf{Kappa}&\textbf{Weighted F1}&\textbf{Bal. Acc.}&\textbf{Kappa}&\textbf{Weighted F1}&\textbf{Bal. Acc.}&\textbf{AUC-PR}&\textbf{AUROC}\\
		\midrule
		EEGNet   
		&58.14 $\pm$ 1.25&44.68 $\pm$ 1.99&57.96 $\pm$ 1.15
		&44.82 $\pm$ 0.94&26.93 $\pm$ 1.21&42.26 $\pm$ 1.08
		&58.89 $\pm$ 1.77&63.11 $\pm$ 1.42&62.83 $\pm$ 1.52\\
		EEGConformer   
		&60.49 $\pm$ 1.04&47.36 $\pm$ 1.71&60.62 $\pm$ 0.95
		&46.96 $\pm$ 1.06&29.24 $\pm$ 1.41&45.33 $\pm$ 1.28
		&59.00 $\pm$ 1.07&63.70 $\pm$ 0.93&63.51 $\pm$ 1.01\\
		\midrule
		LaBraM  
		&61.73 $\pm$ 1.22&49.12 $\pm$ 1.92&61.77 $\pm$ 1.41
		&48.69 $\pm$ 0.85&31.59 $\pm$ 1.54&47.58 $\pm$ 1.03
		&61.66 $\pm$ 1.92&67.61 $\pm$ 0.83&66.04 $\pm$ 0.91\\
		CBraMod  
		&64.17 $\pm$ 0.91&52.22 $\pm$ 1.69&64.27 $\pm$ 1.00
		&51.38 $\pm$ 0.66&35.18 $\pm$ 0.94&49.84 $\pm$ 0.85
		&63.70 $\pm$ 1.51&\underline{71.39} $\pm$ 0.88&69.88 $\pm$ 0.68\\
		CSBrain  
		&63.04 $\pm$ 0.90&50.71 $\pm$ 1.20&63.08 $\pm$ 0.95
		&56.57 $\pm$ 0.71&42.09 $\pm$ 0.93&56.37 $\pm$ 0.87
		&\underline{64.17} $\pm$ 0.37&\textbf{72.30} $\pm$ 0.79&\textbf{72.00} $\pm$ 0.58\\
		REVE
		&\underline{64.80} $\pm$ 1.40&\underline{53.06} $\pm$ 1.87&\underline{64.84} $\pm$ 1.70
		&\textbf{63.96} $\pm$ 0.95&\textbf{51.94} $\pm$ 1.26&\textbf{63.39} $\pm$ 1.10
		&63.84 $\pm$ 0.34&69.90 $\pm$ 0.72&70.63 $\pm$ 0.42\\
		\midrule
		\rrc DeeperBrain  
		&\textbf{65.17} $\pm$ 0.37&\textbf{53.56} $\pm$ 0.49&\textbf{65.38} $\pm$ 0.34
		&\underline{60.69} $\pm$ 1.40&\underline{47.59} $\pm$ 1.86&\underline{60.64} $\pm$ 1.50
		&\textbf{64.31} $\pm$ 0.47&70.99 $\pm$ 0.53&\underline{70.95} $\pm$ 0.43\\
		\midrule
		&\multicolumn{3}{c}{\textbf{ISRUC, 5-class}}&\multicolumn{3}{c}{\textbf{CHB-MIT, 2-class}}&\multicolumn{3}{c}{\textbf{BCIC2020-3, 5-class}}\\
		\cmidrule(lr){2-4}\cmidrule(lr){5-7}\cmidrule(lr){8-10}
		\textbf{Methods}&\textbf{Bal. Acc.}&\textbf{Kappa}&\textbf{Weighted F1}&\textbf{Bal. Acc.}&\textbf{AUC-PR}&\textbf{AUROC}&\textbf{Bal. Acc.}&\textbf{Kappa}&\textbf{Weighted F1}\\
		\midrule
		EEGNet    
		&71.54 $\pm$ 1.21&70.40 $\pm$ 1.73&75.13 $\pm$ 1.24
		&56.58 $\pm$ 1.06&19.14 $\pm$ 1.82&80.48 $\pm$ 1.36
		&44.13 $\pm$ 0.96&30.16 $\pm$ 1.23&44.13 $\pm$ 1.02\\
		EEGConformer   
		&74.00 $\pm$ 1.33&71.43 $\pm$ 1.62&76.34 $\pm$ 1.51
		&59.76 $\pm$ 1.41&22.09 $\pm$ 2.15&82.26 $\pm$ 1.70
		&45.06 $\pm$ 1.33&31.33 $\pm$ 1.83&44.88 $\pm$ 1.54\\
		\midrule
		LaBraM  
		&76.33 $\pm$ 1.02&72.31 $\pm$ 1.82&78.10 $\pm$ 1.33
		&70.75 $\pm$ 3.58&32.87 $\pm$ 4.02&86.79 $\pm$ 1.99
		&50.60 $\pm$ 1.55&38.00 $\pm$ 2.42&50.54 $\pm$ 2.05\\
		CBraMod  
		&78.65 $\pm$ 1.10&74.42 $\pm$ 1.52&\underline{80.11} $\pm$ 0.99
		&\underline{73.98} $\pm$ 2.84&36.89 $\pm$ 3.82&88.92 $\pm$ 1.54
		&53.73 $\pm$ 1.08&42.16 $\pm$ 1.63&53.83 $\pm$ 0.96\\
		CSBrain  
		&\textbf{79.25} $\pm$ 0.30&74.06 $\pm$ 1.02&79.90 $\pm$ 0.91
		&72.62 $\pm$ 1.15&51.64 $\pm$ 4.49&89.15 $\pm$ 3.21
		&56.05 $\pm$ 4.38&45.07 $\pm$ 5.47&56.05 $\pm$ 4.41\\
		REVE
		&78.19 $\pm$ 0.78&\underline{75.00} $\pm$ 1.56&\underline{80.11} $\pm$ 0.99
		&72.02 $\pm$ 4.95&\underline{53.97} $\pm$ 6.38&\underline{90.41} $\pm$ 0.97
		&\underline{56.35} $\pm$ 1.23&\underline{45.43} $\pm$ 1.54&\underline{56.33} $\pm$ 1.24\\
		\midrule
		\rrc DeeperBrain  
		&\underline{78.75} $\pm$ 0.60&\textbf{75.01} $\pm$ 0.32&\textbf{80.89} $\pm$ 0.40
		&\textbf{74.34} $\pm$ 1.49&\textbf{65.51} $\pm$ 4.27&\textbf{94.09} $\pm$ 0.77
		&\textbf{57.06} $\pm$ 0.78&\textbf{46.33} $\pm$ 0.97&\textbf{57.04} $\pm$ 0.78\\
		\midrule
		&\multicolumn{3}{c}{\textbf{MODMA, 2-class}}&\multicolumn{3}{c}{\textbf{SEED-VIG, regression}}&\multicolumn{3}{c}{\textbf{MentalArithmetic, 2-class}}\\
		\cmidrule(lr){2-4}\cmidrule(lr){5-7}\cmidrule(lr){8-10}
		\textbf{Methods}&\textbf{Bal. Acc.}&\textbf{AUC-PR}&\textbf{AUROC}&\textbf{Correlation}&\textbf{R2 Score} & \textbf{RMSE$\downarrow$}&\textbf{Bal. Acc.}&\textbf{AUC-PR}&\textbf{AUROC}\\
		\midrule
		EEGNet    
		&61.90 $\pm$ 2.40&64.50 $\pm$ 3.00&62.30 $\pm$ 4.70
		&57.01 $\pm$ 1.67&23.66 $\pm$ 0.84&28.28 $\pm$ 0.74
		&67.70 $\pm$ 1.16&57.63 $\pm$ 1.02&73.21 $\pm$ 1.08\\
		EEGConformer   
		&63.85 $\pm$ 2.20&66.40 $\pm$ 3.25&64.70 $\pm$ 4.60
		&57.50 $\pm$ 1.39&23.44 $\pm$ 0.91&28.50 $\pm$ 0.83
		&68.05 $\pm$ 1.23&58.29 $\pm$ 1.34&74.24 $\pm$ 1.28\\
		\midrule
		LaBraM  
		&66.81 $\pm$ 7.29&73.43 $\pm$ 5.82&68.83 $\pm$ 6.22
		&59.31 $\pm$ 0.98&24.32 $\pm$ 0.85&27.62 $\pm$ 0.48
		&72.24 $\pm$ 2.05&70.78 $\pm$ 3.77&84.11 $\pm$ 2.46\\
		CBraMod  
		&70.18 $\pm$ 2.34&75.92 $\pm$ 3.05&72.37 $\pm$ 4.91
		&58.49 $\pm$ 2.78&24.82 $\pm$ 3.67&\underline{27.35} $\pm$ 0.83
		&74.39 $\pm$ 2.15&71.68 $\pm$ 2.93&83.90 $\pm$ 3.27\\
		CSBrain  
		&\underline{72.85} $\pm$ 2.60&\underline{77.90} $\pm$ 3.60&\underline{75.10} $\pm$ 5.00
		&\textbf{63.14} $\pm$ 3.56&23.63 $\pm$ 5.19&27.74 $\pm$ 0.94
		&75.58 $\pm$ 1.06&66.96 $\pm$ 2.21&\underline{84.78} $\pm$ 2.97\\
		REVE
		&72.45 $\pm$ 2.10&77.83 $\pm$ 3.21&74.62 $\pm$ 4.75
		&58.77 $\pm$ 1.17&\underline{25.44} $\pm$ 1.90&27.42 $\pm$ 0.35
		&\textbf{76.60} $\pm$ 3.55&\textbf{74.70} $\pm$ 8.07&84.50 $\pm$ 5.14\\
		\midrule
		\rrc DeeperBrain  
		&\textbf{76.31} $\pm$ 2.59&\textbf{81.55} $\pm$ 3.67&\textbf{78.85} $\pm$ 5.19
		&\underline{60.11} $\pm$ 0.72&\textbf{34.14} $\pm$ 1.62&\textbf{25.77} $\pm$ 0.32
		&\underline{75.69} $\pm$ 2.32&\underline{72.04} $\pm$ 4.52&\textbf{86.21} $\pm$ 2.19\\
		\bottomrule
	\end{tabular}\\
	
	\begin{tabular}{c}
		Results are presented as (mean $\pm$ std, \%). \textbf{Bold} denotes the best performance, and \underline{underline} denotes the second best.
	\end{tabular}
	\label{tab:finetuning}
\end{table*}

\section{Experiments}

\subsection{Pretraining Setup}

\subsubsection{Pretraining Datasets and Preprocessing}

DeeperBrain is pretrained on a large-scale and highly diverse EEG corpus comprising 14 public datasets, totaling over 17,200 hours of recordings (2,438,653 non-overlapping samples), as summarized in Table~\ref{tab:pre-train}. This collection spans a wide spectrum of domains, including clinical diagnostics, sleep research, cognitive neuroscience, and perceptual tasks. It exhibits substantial heterogeneity in recording protocols, channel configurations (6 to 128 channels), sampling rates (200–2048 Hz), and epoch durations (5–30 seconds). Such diversity ensures that the model learns representations robust to variations in hardware, montage, and experimental design, a critical requirement for universal EEG decoding.

The corpus includes well-established benchmarks as well as recently released datasets:
\begin{itemize}
	\item \textbf{TUEG}~\citep{obeid2016temple}: A large clinical EEG archive. Due to pervasive artifacts and unmarked noise in the full release, we curated a clean subset of high-quality 30-second segments consistent with CBraMod~\citep{wang2025cbramod}.
	\item \textbf{PhysioNet 2018}~\citep{Goldberger2000PhysioBankPA, ghassemi2018you}: Sleep EEG from 994 subjects (training set only), annotated for sleep staging.
	\item \textbf{Raw EEG Data}~\citep{raweegdata}: Recordings from cognitive task paradigms (information integration and rule-based categorization).
	\item \textbf{Siena Scalp EEG Database}~\citep{Goldberger2000PhysioBankPA, detti2020eeg}: EEG from epilepsy patients, originally recorded with 31 channels but processed here using the standard 10–20 subset of 27 channels.
	\item \textbf{Ten OpenNeuro datasets} (ds006171, ds006317, ds006367, ds006370, ds006437, ds006446, ds006466, ds006480, ds006525, ds006547)~\citep{openneuro}: A collection of contemporary cognitive and clinical EEG studies from the OpenNeuro platform, covering perceptual decision-making, emotional processing, resting-state, and motor imagery paradigms. These datasets vary widely in channel count (28–128) and sampling rate (250–2048 Hz), reflecting modern high-density and multi-center recording practices.
\end{itemize}

All recordings are preprocessed uniformly:  
(i) resampled to 200 Hz (except PhysioNet 2018, which is already at 200 Hz);  
(ii) band-pass filtered (0.3–75 Hz) to remove slow drifts and high-frequency noise;  
(iii) notch-filtered at 50/60 Hz to suppress power-line interference;  
(iv) segmented into non-overlapping fixed-length samples as specified per dataset; and  
(v) normalized to a scale of 100 µV (i.e., divided by 100), so that values predominantly lie in $[-1, 1]$, following the convention in LaBraM~\citep{jiang2024large}.

This extensive and heterogeneous pretraining corpus enables DeeperBrain to internalize both the physical and dynamical principles of EEG, forming the basis for its strong generalization across downstream tasks.

\subsubsection{Pre-training Settings}

DeeperBrain is implemented in Python 3.13.5 using PyTorch 2.8.0 with CUDA 12.9 for GPU acceleration. The model is pretrained on a single machine equipped with an Intel Xeon Gold 6226R CPU and one NVIDIA RTX A5000 GPU, completing training in approximately 7 hours.
The input EEG signal is partitioned into non-overlapping spatiotemporal patches, each corresponding to a 1-second segment (200 time steps at 200 Hz). The backbone architecture consists of a 12-layer Transformer encoder with the following specifications: embedding dimension $D = 200$, feed-forward hidden dimension $D_{\text{ff}} = 800$, and 8 heads of self-attention.
During pretraining, we adopt a random patch-wise masking strategy with a mask ratio of 50\%, where masked patches are replaced by zero vectors in the raw voltage space. The model is trained for 2 epochs with a batch size of 16. We set the coefficients of two pretraining objectives as $\lambda_{\text{MER}} = \lambda_{\text{NSP}} = 1.0$. Optimization is performed using the AdamW optimizer with a base learning rate of $5 \times 10^{-4}$, weight decay of $5 \times 10^{-2}$, and default $\beta_1=0.9$, $\beta_2=0.999$. No learning rate warmup is used, following the pre-normalization design's stability~\citep{wang2019learning}. Gradient clipping with a maximum norm of 1.0 is applied to stabilize training. \textbf{The code and pretrained model will be publicly available.}

\begin{table*}[tb]
	\scriptsize
	\setlength{\tabcolsep}{5pt}
	\centering
	\caption{Performance comparison with existing methods under frozen-probing. }
	\begin{tabular}{lccccccccc} 
		\toprule
		&\multicolumn{3}{c}{\textbf{FACED, 9-class}}&\multicolumn{3}{c}{\textbf{SEED-V, 5-class}}&\multicolumn{3}{c}{\textbf{SEED-VII, 7-class}}\\
		\cmidrule(lr){2-4}\cmidrule(lr){5-7}\cmidrule(lr){8-10}
		\textbf{Methods}&\textbf{Bal. Acc.}&\textbf{Kappa}&\textbf{Weighted F1}&\textbf{Bal. Acc.}&\textbf{Kappa}&\textbf{Weighted F1}&\textbf{Bal. Acc.}&\textbf{Kappa}&\textbf{Weighted F1}\\
		\midrule
		LaBraM  
		& 16.13 $\pm$ 10.04& 5.61 $\pm$ 11.21& 9.88 $\pm$ 13.37
		& \underline{32.63} $\pm$ 0.53& \underline{15.45} $\pm$ 0.69& \underline{32.62} $\pm$ 0.87
		& 15.35 $\pm$ 1.99& 1.29 $\pm$ 2.42& 7.09 $\pm$ 6.15\\
		CBraMod  
		& 25.84 $\pm$ 1.36& 16.32 $\pm$ 1.60& 21.77 $\pm$ 1.46
		& 24.36 $\pm$ 0.38& 5.29 $\pm$ 0.47& 20.39 $\pm$ 1.18
		& \underline{17.32} $\pm$ 0.27& \underline{3.48} $\pm$ 0.35& \underline{14.23} $\pm$ 1.43\\
		CSBrain  
		& 36.59 $\pm$ 0.96& 28.56 $\pm$ 1.07& 36.59 $\pm$ 0.94
		& 20.00 $\pm$ 0.00& 0.00 $\pm$ 0.00& 8.87 $\pm$ 0.00
		& 14.28 $\pm$ 0.01& -0.01 $\pm$ 0.01& 4.76 $\pm$ 0.72\\
		REVE
		& \underline{37.76} $\pm$ 0.54& \underline{29.77} $\pm$ 0.65& \underline{37.63} $\pm$ 0.63
		& 27.82 $\pm$ 0.72& 9.66 $\pm$ 0.87& 27.56 $\pm$ 1.15
		& 14.29 $\pm$ 0.00& 0.00 $\pm$ 0.00& 3.76 $\pm$ 0.42\\
		\midrule
		\rrc DeeperBrain  
		&\textbf{50.96} $\pm$ 0.37& \textbf{44.38} $\pm$ 0.40& \textbf{50.36} $\pm$ 0.36
		&\textbf{35.08} $\pm$ 0.26&\textbf{18.75} $\pm$ 0.29&\textbf{35.39} $\pm$ 0.29
		&\textbf{23.26} $\pm$ 0.22&\textbf{10.71} $\pm$ 0.24&\textbf{22.83} $\pm$ 0.53\\
		\midrule
		&\multicolumn{3}{c}{\textbf{PhysioNet-MI, 4-class}}&\multicolumn{3}{c}{\textbf{BCIC-IV-2a, 4-class}}&\multicolumn{3}{c}{\textbf{SHU-MI, 2-class}}\\
		\cmidrule(lr){2-4}\cmidrule(lr){5-7}\cmidrule(lr){8-10}
		\textbf{Methods}&\textbf{Bal. Acc.}&\textbf{Kappa}&\textbf{Weighted F1}&\textbf{Bal. Acc.}&\textbf{Kappa}&\textbf{Weighted F1}&\textbf{Bal. Acc.}&\textbf{AUC-PR}&\textbf{AUROC}\\
		\midrule
		LaBraM  
		& 46.52 $\pm$ 0.58& 28.70 $\pm$ 0.78& 46.62 $\pm$ 0.45
		& 38.11 $\pm$ 1.04& 17.48 $\pm$ 1.39& 33.50 $\pm$ 1.66
		& \underline{59.48} $\pm$ 3.31& 65.28 $\pm$ 1.21& \textbf{66.92} $\pm$ 1.38\\
		CBraMod  
		& \underline{53.73} $\pm$ 0.39& \underline{38.30} $\pm$ 0.54& \underline{53.91} $\pm$ 0.46
		& 33.06 $\pm$ 0.63& 10.74 $\pm$ 0.84& 21.73 $\pm$ 0.59
		& 57.18 $\pm$ 2.22& 64.96 $\pm$ 0.75& 64.85 $\pm$ 0.49\\
		CSBrain 
		& 26.19 $\pm$ 1.60& 2.04 $\pm$ 2.86& 15.52 $\pm$ 6.61
		& 26.70 $\pm$ 1.05& 2.27 $\pm$ 1.40& 19.29 $\pm$ 3.78
		& 53.05 $\pm$ 0.82& 54.95 $\pm$ 1.78& 54.69 $\pm$ 0.91\\
		REVE
		& 25.01 $\pm$ 0.02& 0.01 $\pm$ 0.03& 10.03 $\pm$ 0.05
		& \underline{42.73} $\pm$ 1.12& \underline{23.63} $\pm$ 1.49& \underline{38.66} $\pm$ 1.31
		& 59.38 $\pm$ 1.14& \underline{66.77} $\pm$ 0.35& 65.63 $\pm$ 0.83\\
		\midrule
		\rrc DeeperBrain
		&\textbf{56.57} $\pm$ 0.73&\textbf{42.10} $\pm$ 0.98&\textbf{56.74} $\pm$ 0.72
		&\textbf{51.01} $\pm$ 2.23&\textbf{34.68} $\pm$ 2.98&\textbf{49.23} $\pm$ 3.09
		&\textbf{60.81} $\pm$ 1.80&\textbf{67.01} $\pm$ 0.27&\underline{66.57} $\pm$ 0.25\\
		\midrule
		&\multicolumn{3}{c}{\textbf{ISRUC, 5-class}}&\multicolumn{3}{c}{\textbf{CHB-MIT, 2-class}}&\multicolumn{3}{c}{\textbf{BCIC2020-3, 5-class}}\\
		\cmidrule(lr){2-4}\cmidrule(lr){5-7}\cmidrule(lr){8-10}
		\textbf{Methods}&\textbf{Bal. Acc.}&\textbf{Kappa}&\textbf{Weighted F1}&\textbf{Bal. Acc.}&\textbf{AUC-PR}&\textbf{AUROC}&\textbf{Bal. Acc.}&\textbf{Kappa}&\textbf{Weighted F1}\\
		\midrule
		LaBraM  
		& 71.79 $\pm$ 0.95& 67.94 $\pm$ 1.08& 74.55 $\pm$ 0.90
		& 55.31 $\pm$ 3.45& 28.76 $\pm$ 3.90& \underline{89.08} $\pm$ 1.95
		& 24.83 $\pm$ 2.61& 6.03 $\pm$ 3.26& 22.76 $\pm$ 5.82\\
		CBraMod  
		& 37.27 $\pm$ 5.90& 24.87 $\pm$ 5.63& 37.16 $\pm$ 3.23
		& 58.93 $\pm$ 2.85& 36.66 $\pm$ 7.10& 86.20 $\pm$ 2.55
		& 20.24 $\pm$ 0.52& 0.30 $\pm$ 0.65& 8.74 $\pm$ 2.12\\
		CSBrain  
		& 22.24 $\pm$ 2.25& 3.01 $\pm$ 3.18& 22.61 $\pm$ 3.11
		& 62.68 $\pm$ 4.42& 8.13 $\pm$ 3.45& 75.48 $\pm$ 4.18
		& 32.19 $\pm$ 1.40& 15.23 $\pm$ 1.75& 30.96 $\pm$ 2.18\\
		REVE
		& \underline{73.55} $\pm$ 0.47& \underline{68.78} $\pm$ 0.27& \underline{75.72} $\pm$ 0.42
		& \underline{63.93} $\pm$ 3.70& \textbf{46.68} $\pm$ 4.06& \textbf{92.11} $\pm$ 1.30
		& \underline{39.04} $\pm$ 9.59& \underline{23.80} $\pm$ 11.99& \underline{36.32} $\pm$ 14.87\\
		\midrule
		\rrc DeeperBrain  
		&\textbf{74.10} $\pm$ 0.58&\textbf{70.07} $\pm$ 0.60&\textbf{76.71} $\pm$ 0.68
		&\textbf{64.95} $\pm$ 2.55&\underline{38.46} $\pm$ 4.84&88.73 $\pm$ 3.75
		&\textbf{43.33} $\pm$ 0.64&\textbf{29.17} $\pm$ 0.80&\textbf{42.69} $\pm$ 0.73\\
		\midrule
		&\multicolumn{3}{c}{\textbf{MODMA, 2-class}}&\multicolumn{3}{c}{\textbf{SEED-VIG, regression}}&\multicolumn{3}{c}{\textbf{MentalArithmetic, 2-class}}\\
		\cmidrule(lr){2-4}\cmidrule(lr){5-7}\cmidrule(lr){8-10}
		\textbf{Methods}&\textbf{Bal. Acc.}&\textbf{AUC-PR}&\textbf{AUROC}&\textbf{Correlation}&\textbf{R2 Score} & \textbf{RMSE$\downarrow$}&\textbf{Bal. Acc.}&\textbf{AUC-PR}&\textbf{AUROC}\\
		\midrule
		LaBraM  
		& 43.12 $\pm$ 7.52& 43.41 $\pm$ 9.28& 30.02 $\pm$ 9.01
		& \underline{56.92} $\pm$ 1.56& \underline{18.51} $\pm$ 4.44& \underline{28.66} $\pm$ 0.80
		& 60.69 $\pm$ 9.18& \underline{69.67} $\pm$ 2.02& \textbf{83.06} $\pm$ 1.94\\
		CBraMod  
		& 46.00 $\pm$ 3.23& 37.55 $\pm$ 1.73& 36.81 $\pm$ 3.62
		& 37.19 $\pm$ 3.20& 9.12 $\pm$ 3.17& 30.27 $\pm$ 0.53
		& 54.79 $\pm$ 1.98& 47.15 $\pm$ 2.70& 65.40 $\pm$ 1.73\\
		CSBrain  
		& 52.38 $\pm$ 10.83& 55.94 $\pm$ 11.95& 52.42 $\pm$ 10.82
		& 22.31 $\pm$ 3.89& 4.38 $\pm$ 1.89& 31.06 $\pm$ 0.31
		& 54.65 $\pm$ 3.77& 38.53 $\pm$ 10.19& 59.32 $\pm$ 6.29\\
		REVE
		& \underline{53.00} $\pm$ 3.48& \underline{61.32} $\pm$ 9.19& \underline{56.75} $\pm$ 9.15
		& 50.32 $\pm$ 4.35& 10.61 $\pm$ 13.46& 29.94 $\pm$ 2.27
		& \underline{67.29} $\pm$ 7.89& \textbf{70.84} $\pm$ 2.13& \underline{82.69} $\pm$ 1.54\\
		\midrule
		\rrc DeeperBrain  
		&\textbf{63.36} $\pm$ 2.40&\textbf{70.37} $\pm$ 1.00&\textbf{70.39} $\pm$ 1.13
		&\textbf{59.85} $\pm$ 1.24&\textbf{31.43} $\pm$ 1.52&\textbf{27.52} $\pm$ 0.30
		&\textbf{68.05} $\pm$ 0.40&54.57 $\pm$ 2.52&74.79 $\pm$ 0.87\\
		\bottomrule
	\end{tabular}\\
	
	\begin{tabular}{c}
		Results are presented as (mean $\pm$ std, \%). \textbf{Bold} denotes the best performance, and \underline{underline} denotes the second best.
	\end{tabular}
	\label{tab:probing}
\end{table*}

\subsection{Downstream Tasks and Evaluation Protocol}

\subsubsection{Downstream Tasks and Datasets}
To comprehensively evaluate the generalization capability of DeeperBrain, we selected 10 diverse downstream BCI tasks. Table~\ref{tab:downstream} summarizes these tasks along with their corresponding datasets. Consistent with the pretraining configuration, all downstream EEG signals were resampled to 200 Hz, and the patch duration was fixed at 1 second (200 data points). To ensure fair comparison, we adopted standard data splitting protocols from LaBraM~\citep{jiang2024large} or CBraMod~\citep{wang2025cbramod} whenever available.

\subsubsection{Fine-tuning and Frozen-probing Settings}

For \textit{end-to-end fine-tuning}, the model is optimized using the AdamW optimizer with a weight decay of $5 \times 10^{-2}$. The initial learning rate is selected from $\{1 \times 10^{-4}, 3 \times 1 0^{-4}, 5 \times 10^{-4}\}$ based on validation performance, and is decayed over the course of training via cosine annealing. Training proceeds for 50 epochs with a batch size of either 32 or 64, depending on dataset scale and memory constraints. In the \textit{frozen-probing} setting, the pretrained DeeperBrain backbone is kept entirely frozen, and only the downstream prediction head is trained. The same optimization protocol such as optimizer, learning rate schedule, batch size, and number of epochs, is applied to ensure a fair comparison between the two adaptation strategies.

\subsubsection{Compared Methods}
We compare DeeperBrain with both non-foundation-model and foundation-model baselines on all the downstream BCI tasks.
We adopt \textbf{EEGNet}~\citep{lawhern2018eegnet} and \textbf{EEGConformer}~\citep{song2022eeg} as non-foundation-model baselines.
Concurrently, we use four very strong existing methods, \textbf{LaBraM}~\citep{jiang2024large}, \textbf{CBraMod}~\citep{wang2025cbramod}, \textbf{CSBrain}~\citep{zhou2025csbrain} and \textbf{REVE}~\citep{ouahidi2025reve}  as the foundation-model baselines.

\subsubsection{Metrics}
We employ specific evaluation protocols based on the task type. For \textit{binary classification}, we report \textbf{Balanced Accuracy}, \textbf{AUC-PR}, and \textbf{AUROC}, with AUROC serving as the primary monitor score. For \textit{multi-class classification}, we utilize \textbf{Balanced Accuracy}, \textbf{Cohen’s Kappa}, and \textbf{Weighted F1}, selecting Cohen’s Kappa as the monitor score. For \textit{regression} tasks, performance is assessed via \textbf{Pearson’s Correlation}, \textbf{R2 Score}, and \textbf{RMSE}, with the R2 score used for monitoring. All reported results represent the mean and standard deviation across five independent random seeds.

\subsection{Performance Comparison with End-to-End Fine-tuning}

We evaluate the effectiveness of DeeperBrain under the standard \textit{end-to-end fine-tuning} protocol. As shown in Table~\ref{tab:finetuning}, DeeperBrain achieves state-of-the-art or highly competitive performance across a comprehensive suite of EEG decoding benchmarks. The extensive experimental results reveal three key insights regarding the advantages of our methodology.

First, DeeperBrain consistently outperforms traditional supervised baselines, such as EEGNet and EEGConformer, by a significant margin across all datasets. This validates the efficacy of large-scale self-supervised pretraining in extracting generalizable representations, effectively overcoming the generalization bottlenecks inherent in training deep networks on limited task-specific data.

Second, compared to existing foundation models that rely on general-purpose sequence modeling, DeeperBrain demonstrates superior transferability. While purely data-driven approaches implicitly infer structure from vast corpora, our model achieves leading performance by explicitly incorporating domain-specific inductive biases. The consistent superiority across diverse paradigms suggests that the Neurodynamics Statistics Prediction (NSP) objective captures fundamental properties of neural computation. Unlike reconstruction-only objectives that prioritize local waveform fidelity, NSP compels the model to internalize macroscopic order parameters, including spectral organization and functional integration. These dynamical features function as a universal language for brain activity, facilitating effective decoding across distinct cognitive and pathological states.

Third, the robustness observed across heterogeneous datasets highlights the critical utility of our neurophysiologically grounded positional encodings. By modeling volume conduction via 3D electrode geometry, DeeperBrain effectively mitigates spatial domain shifts caused by varying electrode montages. Concurrently, the temporal encoding, utilizing oscillatory and decay bases, captures slow neural modulations and adaptation dynamics. This capability enables the model to generalize effectively across tasks with distinct temporal characteristics, regardless of whether the target features are transient events or continuous state fluctuations. These results substantiate our hypothesis that aligning deep learning architectures with the biophysical and dynamical first principles of neuroscience is essential for constructing a truly universal and robust EEG foundation model.

\begin{figure*}[tb]
	\centering
	\small
	\begin{minipage}[t]{\textwidth}
		\centering
		\includegraphics[width=1.0\textwidth]{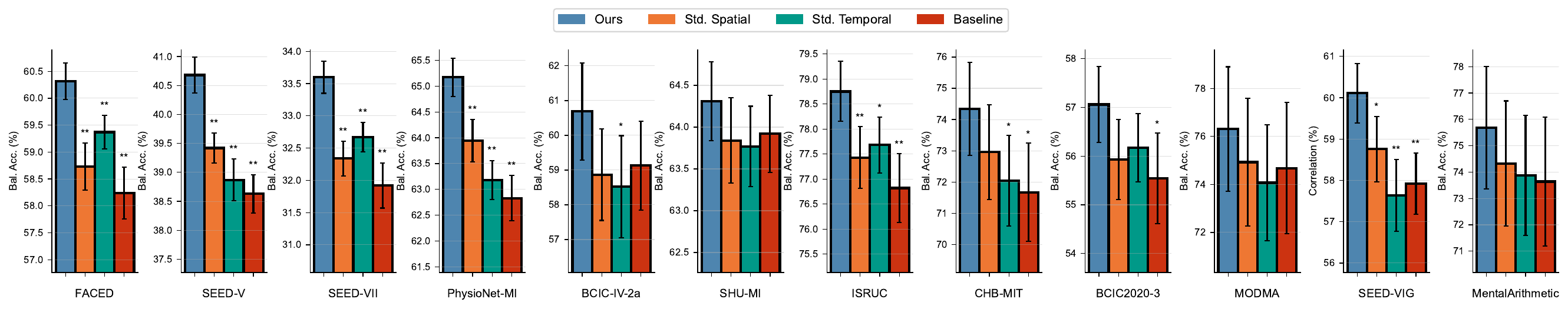}
		\scriptsize 
		\centering
		\begin{tabular}{c}
			(a) \textit{end-to-end fine-tuning}
		\end{tabular}
	\end{minipage}
	\begin{minipage}[t]{\textwidth}
		\centering
		\includegraphics[width=1.0\textwidth]{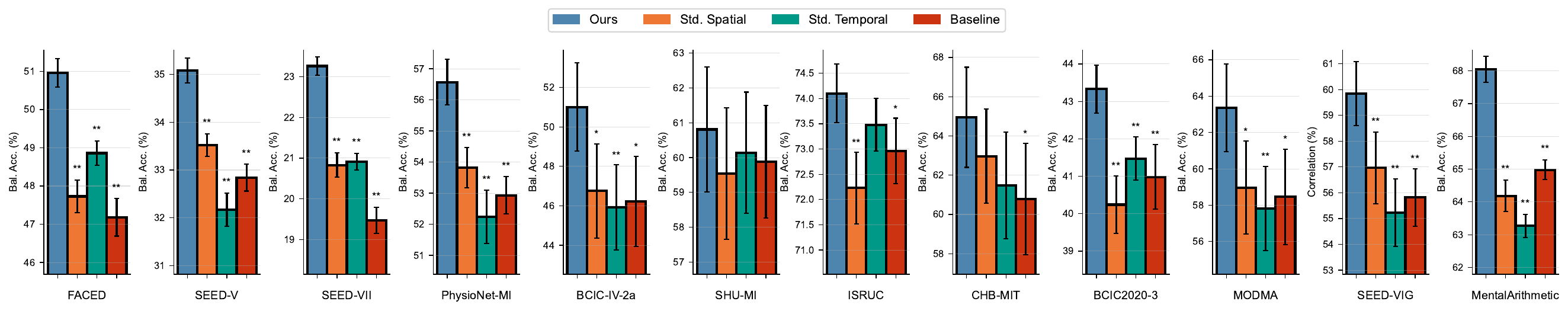}
		\scriptsize 
		\centering
		\begin{tabular}{c}
			(b) \textit{frozen-probing}
		\end{tabular}
	\end{minipage} 
	\caption{Performance comparison (mean $\pm$ std, \%) in balanced accuracy on positional encoding. ``*'' indicates $p<0.05$ and ``**'' indicates $p<0.01$.}
	\label{fig:ablation_pos_enc}
\end{figure*}

\subsection{Performance Comparison with Frozen-probing}

To rigorously assess the intrinsic generalization capability of the learned representations, we evaluate DeeperBrain under a \textit{frozen-probing} protocol. In this setting, the pretrained backbone remains frozen, and only a task-specific prediction head is trained for each downstream dataset. As reported in Table~\ref{tab:probing}, DeeperBrain achieves dominant performance across the majority of benchmarks, significantly outperforming existing foundation models that often struggle to generalize without extensive parameter adaptation.

The experimental outcomes provide compelling evidence for the semantic richness of our neuro-grounded representations. Unlike \textit{end-to-end fine-tuning}, which allows the model to re-optimize its parameters for specific target distributions, \textit{frozen-probing} tests whether the pretrained features are separable and universally applicable "out-of-the-box." The results indicate that DeeperBrain does not merely memorize statistical patterns but successfully encapsulates core neurodynamical properties, such as oscillatory states and functional connectivity, that are invariant across different subjects and paradigms. Consequently, the frozen features alone are sufficient to support high-performance decoding in complex tasks ranging from emotion recognition to mental workload estimation.

Notably, while fine-tuning generally yields the highest absolute performance by adapting to task-specific nuances, the performance drop observed when switching to \textit{frozen-probing} is remarkably smaller for DeeperBrain compared to baselines. This resilience suggests that our dual-objective pretraining strategy, which combines signal reconstruction with neurodynamics prediction, effectively forces the model to disentangle meaningful physiological factors from noise during the pretraining phase itself. This characteristic is particularly valuable for practical brain-computer interfaces, as it implies that DeeperBrain can serve as a robust, distinct feature extractor for new tasks with minimal computational overhead and reduced risk of overfitting to small calibration datasets.

\begin{figure*}[tb]
	\centering
	\small
	\begin{minipage}[t]{\textwidth}
		\centering
		\includegraphics[width=1.0\textwidth]{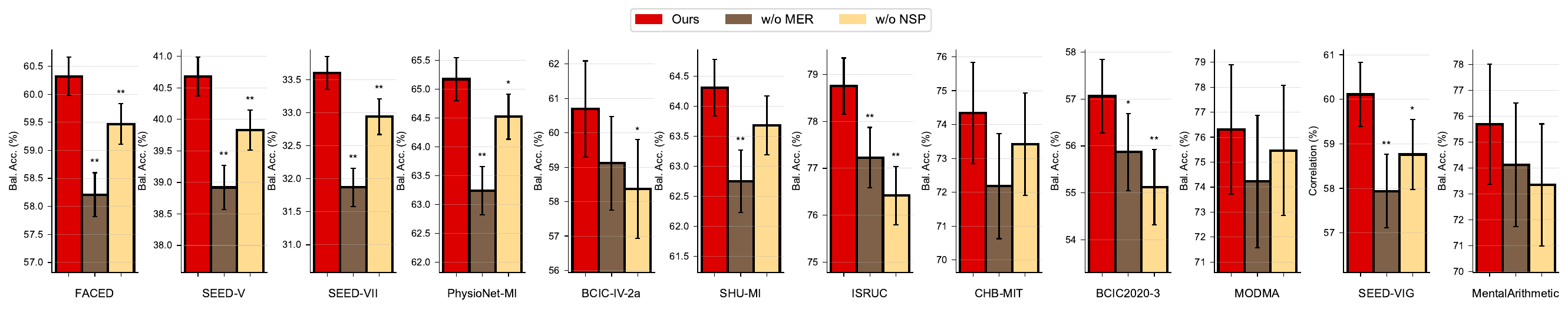}
		\scriptsize 
		\centering
		\begin{tabular}{c}
			(a) \textit{end-to-end fine-tuning}
		\end{tabular}
	\end{minipage}
	\begin{minipage}[t]{\textwidth}
		\centering
		\includegraphics[width=1.0\textwidth]{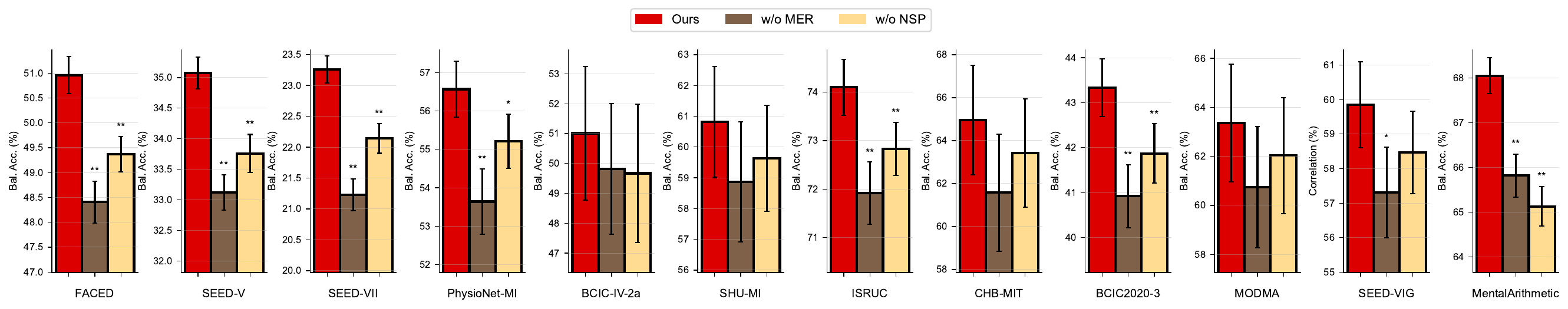}
		\scriptsize 
		\centering
		\begin{tabular}{c}
			(b) \textit{frozen-probing}
		\end{tabular}
	\end{minipage} 
	\caption{Performance comparison (mean $\pm$ std, \%) in balanced accuracy on pretraining objective. ``*'' indicates $p<0.05$ and ``**'' indicates $p<0.01$.}
	\label{fig:ablation_objective}
\end{figure*}

\subsection{Positional Encoding Comparison}

To rigorously assess the specific contribution of our neurophysiologically grounded designs against standard deep learning components, we conducted a controlled ablation study. We compare the full \textbf{DeeperBrain} model against three variants: 
(1) \textbf{Std. Spatial}, where our volume-conduction-aware encoding is replaced by a standard linear projection of 3D electrode coordinates; 
(2) \textbf{Std. Temporal}, where our dynamic temporal basis is replaced by standard sinusoidal positional embeddings; and 
(3) \textbf{Baseline}, which utilizes both standard spatial and temporal embeddings, effectively stripping the model of specific neurophysical priors. 
Figure~\ref{fig:ablation_pos_enc} illustrates the comparative performance across twelve diverse datasets.

The empirical results consistently demonstrate that DeeperBrain outperforms the \textit{Std. Spatial} variant. This finding indicates that merely providing geometric coordinates is insufficient. By explicitly modeling volume conduction through a distance-decay kernel, DeeperBrain provides a strong inductive bias reflecting the physical mixing of electrical fields, enabling the Transformer to better disentangle neural sources from blurred sensor observations.
Furthermore, the superiority of DeeperBrain over the \textit{Std. Temporal} variant highlights the limitation of generic sinusoidal positional embeddings, which capture sequence order but lack the intrinsic oscillatory and adaptive structure of brain dynamics. Our method aligns the model's temporal representation with biological timescales. Finally, the significant gap between DeeperBrain and the \textit{Baseline} confirms that these spatial and temporal priors provide complementary benefits. Collectively, these results substantiate that embedding EEG-specific physics into the architecture yields representations significantly more robust than those derived from generic sequence modeling components.

\subsection{Pretraining Objective Comparison}

To evaluate the individual contributions of our dual-objective strategy, we performed an ablation study by training DeeperBrain with only Masked EEG Reconstruction (w/o NSP) and only Neurodynamics Statistics Prediction (w/o MER). Figure~\ref{fig:ablation_objective} summarizes the performance under both \textit{end-to-end fine-tuning} and \textit{frozen-probing} protocols.

The results consistently show that the full DeeperBrain model, which jointly optimizes both objectives, achieves the highest performance. This superiority indicates that signal fidelity and dynamical consistency are not redundant but rather complementary goals in EEG representation learning. Specifically, the performance degradation observed in the "w/o MER" variant suggests that reconstruction is essential for capturing fine-grained morphological details and high-frequency transients. Without the constraint to recover the original waveform, the model may overlook subtle temporal features that are critical for tasks such as seizure detection.
Conversely, the decline in performance for the "w/o NSP" variant highlights the necessity of explicitly enforcing neurophysiological alignment. While reconstruction ensures that the model encodes the "what" of the signal, the NSP objective guides the model to understand the "how" and "why" by predicting macroscopic order parameters like spectral power distribution and functional connectivity. This mechanism encourages the model to abstract away from raw noise and focus on the emergent dynamical structures that define brain states. The robustness of the full model, particularly in the probing setting, underscores that combining these two perspectives enables DeeperBrain to learn a universal representation that is both physically accurate in the time domain and mechanistically meaningful in the neurodynamical domain.

\begin{figure}[tb]
	\centering
	\small
	\includegraphics[width=\columnwidth]{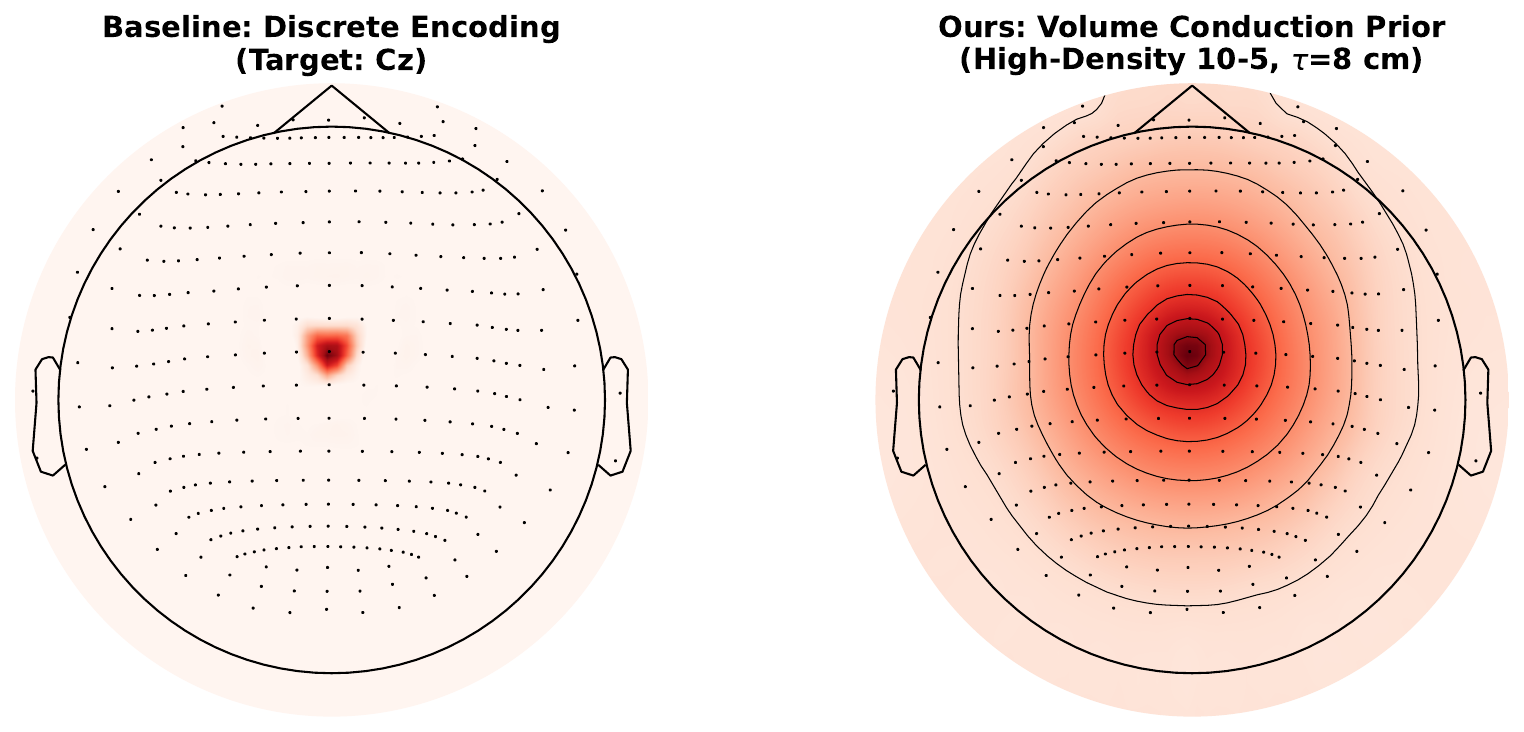}
	\caption{Comparison of spatial receptive fields on a high-density 10-5 montage. Unlike the baseline (left) which exhibits a discrete impulse response, DeeperBrain (right) explicitly models \textit{volume conduction} via a continuous spatial decay kernel (visualized here with $\tau=8$ cm), aligning with the physics of scalp potentials.}
	\label{fig:vis_spatial}
\end{figure}

\begin{figure*}[tb]
	\centering
	\small
	\includegraphics[width=0.99\textwidth]{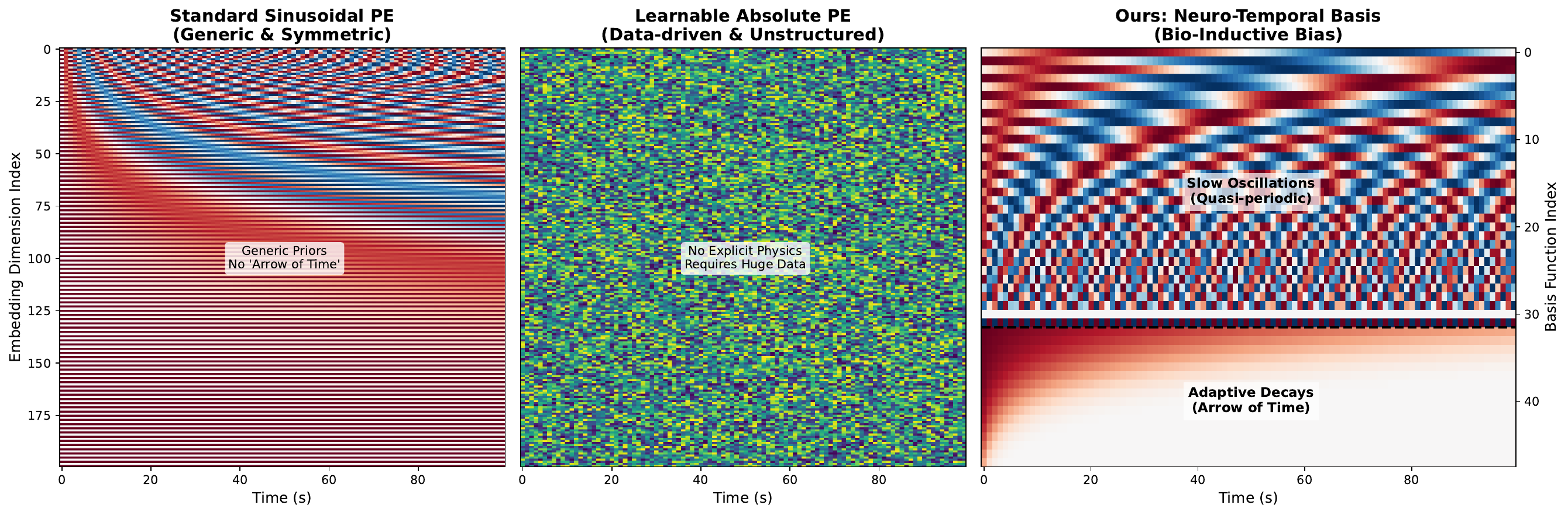}
	\caption{Comparison of temporal positional encodings. \textbf{(Left)} Standard sinusoidal PE relies on generic, symmetric spectral priors. \textbf{(Middle)} Learnable absolute PE starts as unstructured noise, lacking explicit physical constraints. \textbf{(Right)} DeeperBrain PE incorporates strong neurophysiological inductive biases: \textit{Slow Oscillations} for quasi-periodic state maintenance and \textit{Adaptive Decays} for modeling the dissipative "arrow of time".}
	\label{fig:vis_temporal}
\end{figure*}

\begin{figure*}[tb]
	\centering
	\small
	\includegraphics[width=0.99\textwidth]{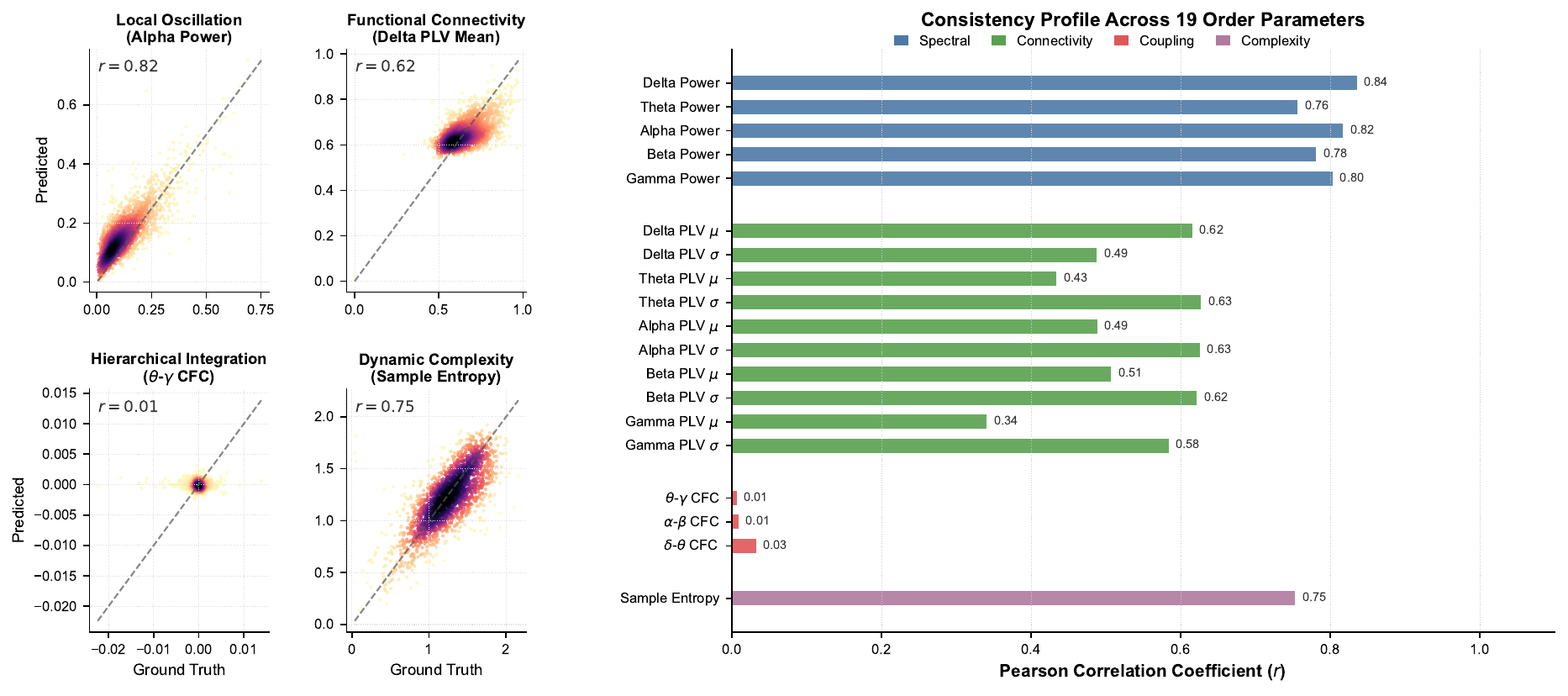}
	\caption{Visualization of Neurodynamical Consistency (Zero-shot on FACED). \textbf{(Left)} Scatter density plots compare ground-truth vs. predicted statistics. Strong alignment is observed for robust metrics like Alpha Power ($r=0.82$) and Sample Entropy ($r=0.75$). \textbf{(Right)} The consistency profile across 19 metrics confirms high performance for spectral and complexity measures. The low correlation for CFC reflects the sparsity of this feature in the target task and the model's robustness against fitting noise.}
	\label{fig:vis_consistency}
\end{figure*}

\subsection{Visualization Analysis}
\label{subsec:vis_analysis}

While quantitative benchmarks demonstrate DeeperBrain's superior performance, it is equally crucial to verify that these gains stem from genuine modeling of neurophysiological mechanisms rather than statistical overfitting. In this section, we qualitatively analyze the learned spatial and temporal structures, as well as the neurodynamical consistency of the optimization objective.

\subsubsection{Modeling Volume Conduction in Channel Positional Encoding}
Standard EEG modeling typically treats electrodes as discrete, isolated nodes, neglecting the biophysical reality of \textit{volume conduction}—the passive spread of electrical currents through head tissues. 

To validate DeeperBrain's ability to capture this physical phenomenon, we visualize the effective spatial receptive field of a central electrode (Cz) in Fig.~\ref{fig:vis_spatial}. The baseline (left) displays a discrete "impulse" response, implying an assumption of sensor independence that contradicts physical laws. In contrast, DeeperBrain (right) generates a smooth, isotropic diffusion pattern. By explicitly modeling distance-dependent decay (shown with a physical prior $\tau \approx 8$ cm), DeeperBrain acts as a spatial regularizer. This design encourages the Transformer to aggregate information from a physically meaningful neighborhood, effectively disentangling the underlying neural source from the volume-conducted observation.

\subsubsection{Oscillations and Adaptation in Temporal Positional Encoding}
Neural dynamics span multiple timescales, ranging from millisecond-level spiking to minute-level state transitions. However, generic positional encodings, including standard sinusoidal and learnable embeddings, treat time indices as symmetric dimensions or arbitrary markers. They lack inductive biases for the "arrow of time," failing to capture the dissipative nature of neural adaptation.

Fig.~\ref{fig:vis_temporal} compares the feature maps of three positional encoding strategies over a 100-second window. The standard encoding (left) relies on generic spectral priors without directionality. The learnable encoding (middle) exhibits an unstructured, high-entropy distribution, implying that the model must learn temporal rules entirely from data, which risks overfitting in data-scarce regimes. In contrast, DeeperBrain's basis (right) reveals a structured, disentangled representation aligned with neural dynamics:
\begin{itemize}
	\item \textbf{Slow Oscillations (Top):} The upper dimensions encode stable, quasi-periodic structures (periods 2s--100s), providing the necessary context to identify macroscopic rhythms.
	\item \textbf{Adaptive Decays (Bottom):} The lower dimensions exhibit a smooth, directional fading gradient. Unlike symmetric sinusoids or random embeddings, these decay bases explicitly model the "arrow of time," capturing \textit{neural adaptation} and the gradual forgetting of past context.
\end{itemize}

\subsubsection{Neurodynamical Consistency of Optimization Objective}
\label{subsec:nsp_vis}

Beyond local reconstruction, the Neurodynamics Statistics Prediction (NSP) objective forces the model to internalize macroscopic order parameters. To verify this semantic alignment, we evaluate DeeperBrain on the downstream FACED dataset in a \textit{zero-shot} manner, comparing ground-truth statistics (calculated from full signals) against predictions inferred from 50\% masked inputs.

Fig.~\ref{fig:vis_consistency} presents the consistency profile across 19 neurodynamical metrics. DeeperBrain exhibits remarkable consistency for robust physiological markers. Spectral power (e.g., Alpha band $r=0.82$) and dynamic complexity (Sample Entropy $r=0.75$) show strong linear alignment, indicating that the model successfully identifies oscillatory states and signal regularity even from partial observations. Functional connectivity (PLV) also maintains moderate consistency ($r \approx 0.6$), suggesting the capture of global synchronization patterns.

Notably, Cross-Frequency Coupling (CFC) exhibits lower correlation ($r \approx 0.01$). This discrepancy offers a valuable neurophysiological insight: (1) \textbf{Sparsity:} CFC is a high-order non-linear statistic that is typically sparse and low-SNR in scalp EEG during passive viewing tasks (like FACED); (2) \textbf{Phase Uncertainty:} Recovering precise instantaneous phase relationships under high masking ratios is an ill-posed problem. The model's conservative prediction on CFC suggests it prioritizes robust, salient features (e.g., power and entropy) over fitting spurious correlations in noise-dominated metrics.

\section{Discussion}
\label{sec:discussion}

\subsection{Implications}
DeeperBrain advances EEG foundation modeling by bridging the gap between deep learning and established neurophysiological principles. Unlike prior approaches that rely on general-purpose sequence modelers to implicitly infer neural patterns, our framework explicitly integrates volume conduction-aware spatial priors and multi-scale temporal dynamics. These inductive biases align the model structure with the physical and functional organization of the brain. By transcending purely data-driven feature extraction, this principled approach enhances performance across diverse downstream tasks and improves neuroscientific interpretability, ensuring that learned representations intrinsically encode canonical markers including spectral organization, functional connectivity, cross-frequency coordination, and dynamic complexity.

From an engineering perspective, DeeperBrain demonstrates that embedding domain knowledge into foundation models yields superior transferability. The robust performance observed under both \textit{end-to-end fine-tuning} and \textit{frozen-probing} protocols indicates that neurodynamics-guided self-supervision is more effective than reconstruction alone. This finding holds profound implications for the deployment of universal BCIs where labeled data is scarce and inter-subject variability is high. Consequently, a single pretrained DeeperBrain model can serve as a universal encoder for a wide range of BCI tasks, requiring minimal task-specific adaptation and effectively lowering the barrier for real-world application.

\subsection{Limitations and Future Directions}
Despite these advancements, we acknowledge specific limitations in the current framework. First, DeeperBrain relies on standardized 3D electrode coordinates which may be absent in some clinical settings. Future iterations could integrate coordinate estimation or learned spatial priors to address geometric uncertainty. Second, the NSP objective currently utilizes four classical statistics. Expanding this set to higher-order dynamics, such as metastability and criticality, may further align representations with complex brain states. Third, while stochastic reorganization accommodates variable channel counts, explicitly modeling montage topology shifts warrants further exploration to enhance robustness.

Furthermore, although our pretraining corpus spans 14 heterogeneous datasets, it retains demographic biases toward specific populations due to the geographic concentration of data sources. Extending DeeperBrain to globally representative EEG data would enhance fairness and generalizability. Finally, the current reliance on fixed-duration patches limits flexibility. Future work will explore streaming or variable-length inference mechanisms to facilitate real-time BCI applications.

\section{Conclusion}
\label{sec:conclusion}
We have presented DeeperBrain, a neuro-grounded EEG foundation model designed to facilitate universal BCI. By explicitly integrating biophysical inductive biases, specifically spatial volume conduction and temporal multi-scale adaptation, and enforcing global dynamical consistency via neurodynamics statistics prediction, DeeperBrain acquires representations that are both physically grounded and functionally meaningful. Empowered by pretraining on over 17,000 hours of diverse unlabeled EEG, the model achieves state-of-the-art or highly competitive performance across a broad spectrum of downstream tasks, notably maintaining superior efficacy even under a rigorous \textit{frozen-probing} protocol. These results demonstrate that aligning machine learning architectures with neuroscientific first principles yields significant empirical gains. We envision DeeperBrain as a pivotal step toward universal BCI, marking a paradigm shift from purely data-driven empiricism to principled, neuro-grounded intelligence that fundamentally understands the language of the brain.

\begin{footnotesize}
	\bibliographystyle{IEEEtranN}
	\bibliography{reference}
\end{footnotesize}

\section{Biography Section}
\begin{IEEEbiography}[{\includegraphics[width=1in,height=1.25in,clip,keepaspectratio]{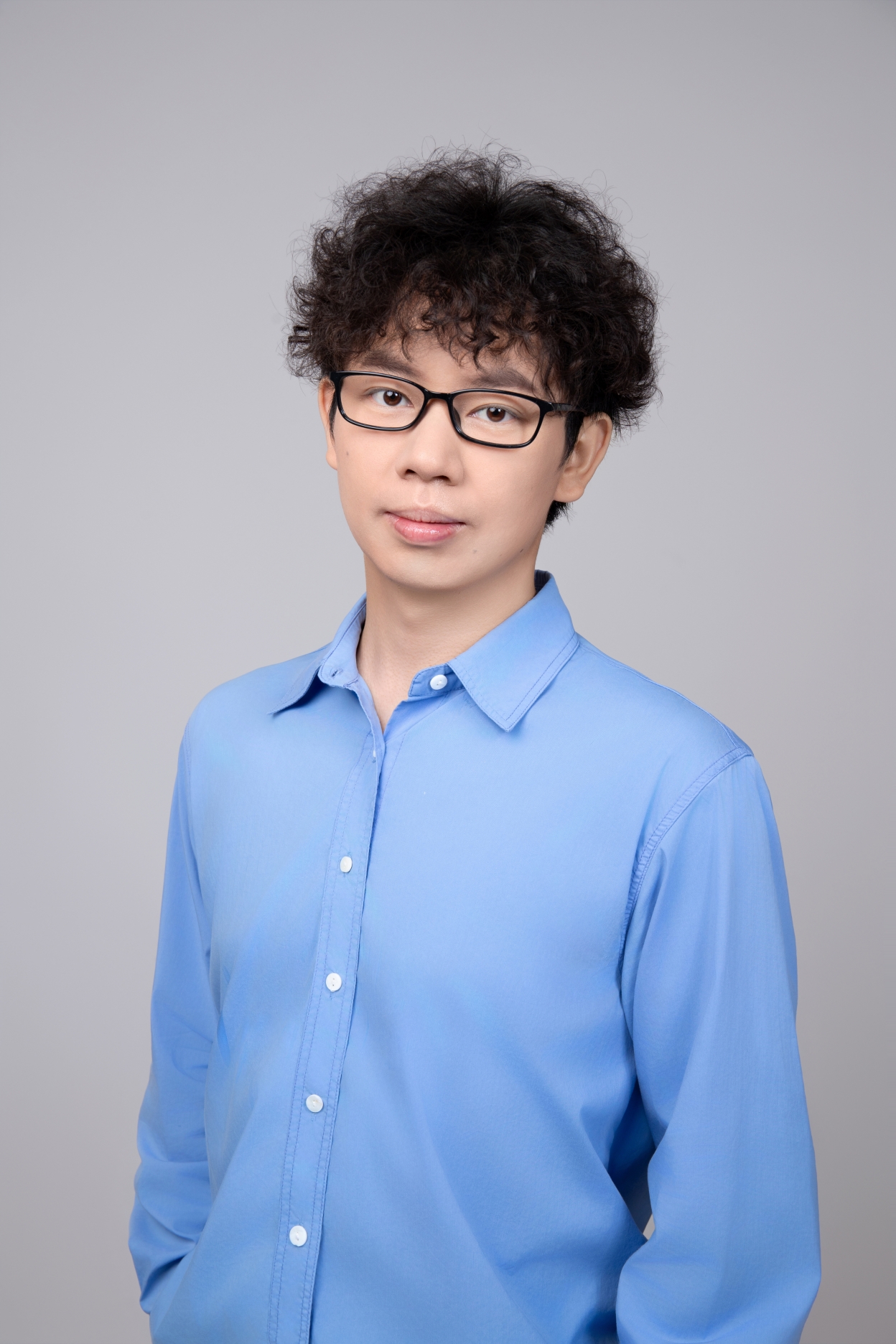}}]{Jiquan Wang}
	received the Ph.D. degree in Computer Science and Technology from Zhejiang University, Hangzhou, China, in 2025.
	
	He is currently a research fellow at the State Key Laboratory of Brain-Machine Intelligence, Zhejiang University. His research interests include EEG decoding, brain-computer interfaces and artificial intelligence. He received the 2025 ACM Hangzhou Outstanding Doctoral Dissertation Award.
\end{IEEEbiography}

\begin{IEEEbiography}[{\includegraphics[width=1in,height=1.25in,clip,keepaspectratio]{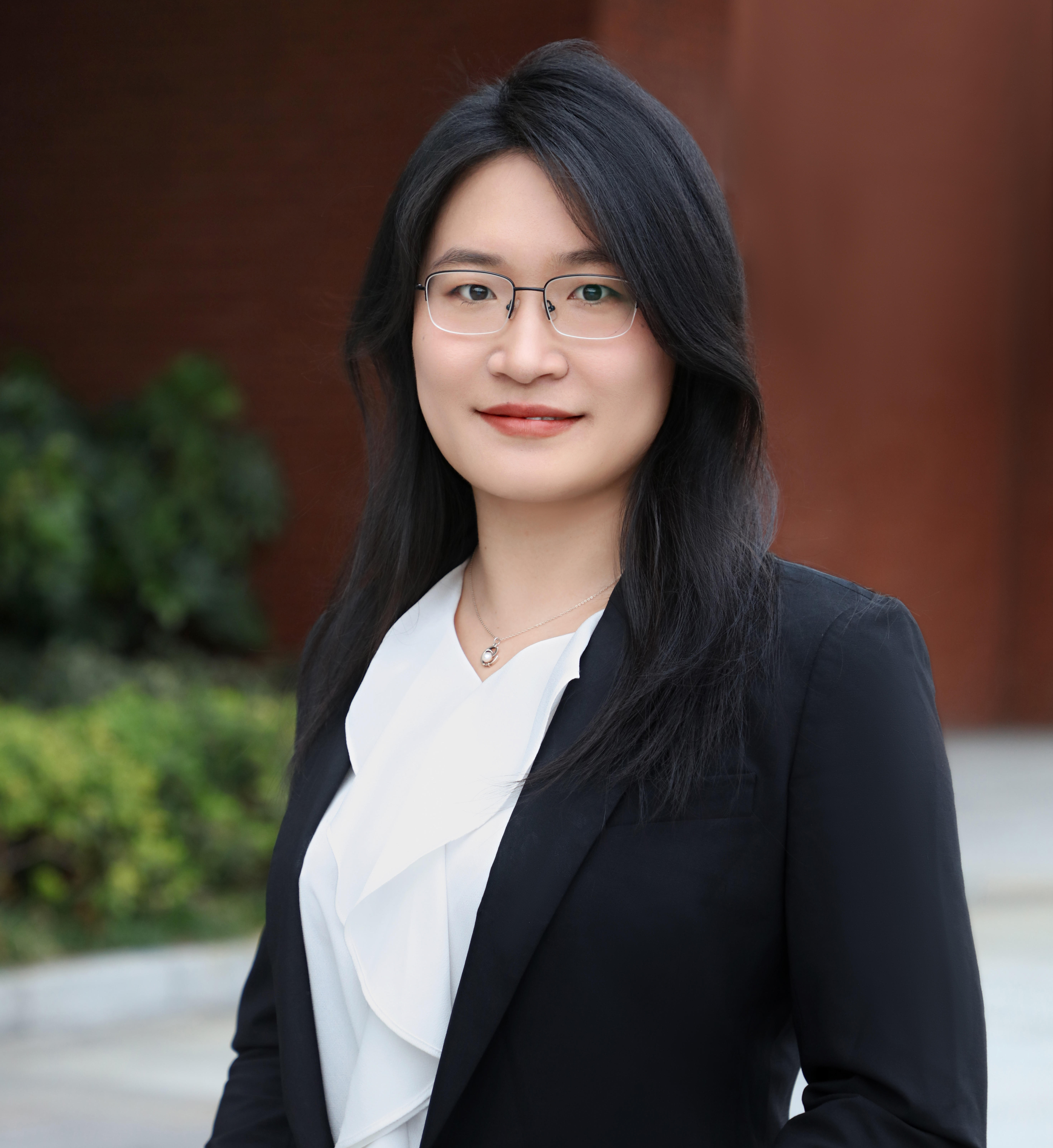}}]{Sha Zhao}
	received the Ph.D. degree from Zhejiang
	University, Hangzhou, China, in 2017. 
	
	She is currently a research professor with the College of Computer Science and Technology, Zhejiang University. She visited the Human-Computer Interaction Institute, Carnegie Mellon University, Pittsburgh, PA, USA, as a visiting Ph.D. student from 2015 to 2016. Her research interests include brain-machine interfaces, data mining and machine learning. 
	Dr. Zhao received the Best Paper Award of ACM UbiComp’16.
\end{IEEEbiography}

\begin{IEEEbiography}[{\includegraphics[width=1in,height=1.25in,clip,keepaspectratio]{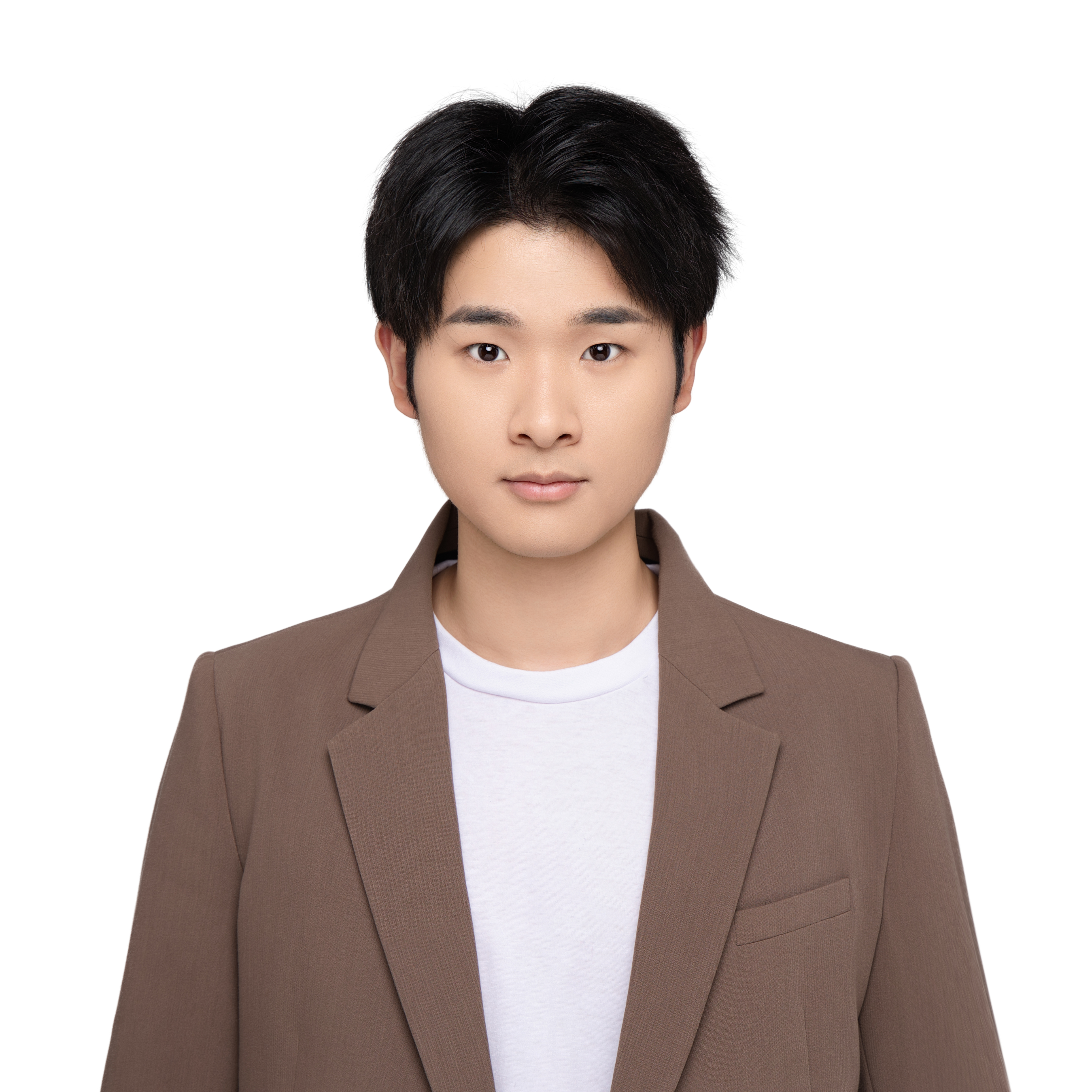}}]{Yangxuan Zhou}
	received the B.S. degree in Electronic Information from Shanghai University, Shanghai, China, in 2022.
	
	He is currently working toward the Ph.D. degree of Computer Science and Technology with Zhejiang University, Hangzhou, China. His research interests include EEG decoding, Brain-Computer Interfaces, Deep Learning, and Artificial Intelligence.
\end{IEEEbiography}

\begin{IEEEbiography}[{\includegraphics[width=1in,height=1.25in,clip,keepaspectratio]{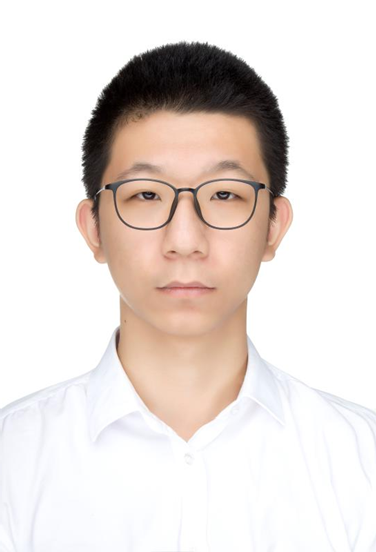}}]{Yiming Kang}
	received the B.S. degree in Artificial Intelligence from Soochow University, Suzhou, China, in 2024.
	
	He has been pursuing the M.S. degree in Electronic Information with Zhejiang University, Hangzhou, China, since 2025. His research primarily focuses on multimodal brain foundation model in the field of non-invasive brain-computer interfaces (BCI).
\end{IEEEbiography}

\begin{IEEEbiography}[{\includegraphics[width=1in,height=1.25in,clip,keepaspectratio]{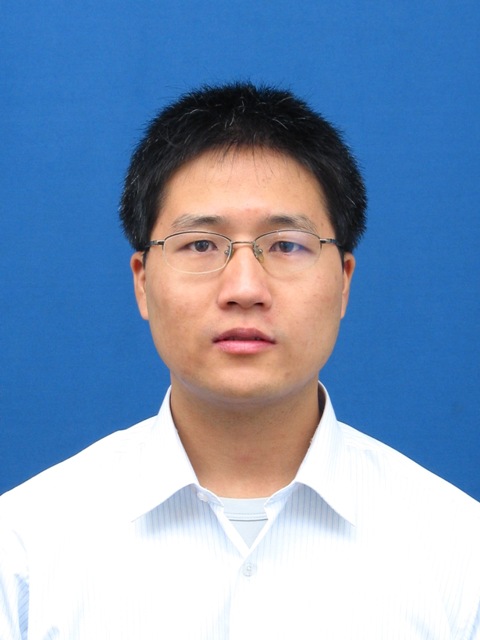}}]{Shijian Li}
	received the Ph.D. degree from Zhejiang
	University, Hangzhou, China, in 2006.
	
	In 2010, he was a Visiting Scholar with the
	Institute Telecom SudParis, Évry, France. He is currently with the College of Computer Science and Technology, Zhejiang University. He has published over 40 papers. His research interests include sensor networks, ubiquitous computing, and social computing.
	Dr. Li serves as an Editor for the International Journal of Distributed Sensor Networks and as a reviewer or the PC member of over ten conferences.
\end{IEEEbiography}

\begin{IEEEbiography}[{\includegraphics[width=1in,height=1.25in,clip,keepaspectratio]{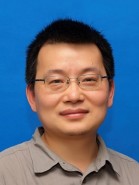}}]{Gang Pan} 
	(Senior Member, IEEE) received the B.Eng. and Ph.D. degrees from Zhejiang University, Hangzhou, China, in 1998 and 2004, respectively.
	
	He is currently a Professor with the Department of Computer Science, and the Director of the State Key Laboratory of Brain-Machine Intelligence, Zhejiang University. From 2007 to 2008, he was a Visiting Scholar with the University of California, Los Angeles, CA, USA. His current interests include brain-inspired computing, brain-machine interfaces and artificial intelligence.
\end{IEEEbiography}

\vfill

\end{document}